\documentclass{article}

\usepackage{microtype}
\usepackage{graphicx}
\usepackage{subfigure}
\usepackage{booktabs} 

\usepackage{amsmath,amsfonts,bm}

\def\eqref#1{equation~\ref{#1}}

\def\1{\bm{1}}

\DeclareMathAlphabet{\mathsfit}{\encodingdefault}{\sfdefault}{m}{sl}
\SetMathAlphabet{\mathsfit}{bold}{\encodingdefault}{\sfdefault}{bx}{n}

\usepackage{hyperref}
\usepackage{authblk}
\usepackage{siunitx}

\usepackage{appendix}

\usepackage{float}

\usepackage[accepted]{icml2021}

\usepackage{selectp}

\icmltitlerunning{Perceiver: General Perception with Iterative Attention}

\begin{document}

\newcommand{\mycomment}[3]{{\textcolor{#3}{#1 #2}}}
\newcommand{\drewmarker}[1]{\textcolor{blue}{\emph{Drew:} #1}}
\newcommand{\tododrew}[1]{\textcolor{blue}{\emph{TODO(Drew):} #1}}

\newcommand{\joaomarker}[1]{\textcolor{blue}{\emph{Joao:} #1}}
\newcommand{\todojoao}[1]{\textcolor{blue}{\emph{TODO(Joao):} #1}}

\newcommand{\felixmarker}[1]{\textcolor{magenta}{\emph{Felix:} #1}}
\newcommand{\todofelix}[1]{\textcolor{magenta}{\emph{TODO(Felix):} #1}}

\definecolor{orange}{RGB}{244, 122, 47}
\definecolor{my_green}{RGB}{74, 184, 72}
\definecolor{my_blue}{RGB}{23, 117, 187}

\twocolumn[
\icmltitle{Perceiver: General Perception with Iterative Attention}

\icmlsetsymbol{equal}{*}

\begin{icmlauthorlist}
\icmlauthor{Andrew Jaegle}{dm}
\icmlauthor{Felix Gimeno}{dm}
\icmlauthor{Andrew Brock}{dm}
\icmlauthor{Andrew Zisserman}{dm}
\icmlauthor{Oriol Vinyals}{dm}
\icmlauthor{Joao Carreira}{dm}
\end{icmlauthorlist}

\icmlaffiliation{dm}{DeepMind -- London, UK}

\icmlcorrespondingauthor{Andrew Jaegle}{drewjaegle@deepmind.com}

\icmlkeywords{Perceiver, Transformer, Attention, Cross-attention, Image Transformers, Vision Transformer, Multimodal, ImageNet, Permuted ImageNet, AudioSet, ModelNet}

\vskip 0.3in
]



\printAffiliationsAndNotice{}

\begin{abstract}
Biological systems perceive the world by simultaneously processing high-dimensional inputs from modalities as diverse as vision, audition, touch, proprioception, etc.
The perception models used in deep learning on the other hand are designed for individual modalities, often relying on domain-specific assumptions such as the local grid structures exploited by virtually all existing vision models. These priors introduce helpful inductive biases, but also lock models to individual modalities.
In this paper we introduce \textit{the Perceiver} -- a model that builds upon Transformers and hence makes few architectural assumptions about the relationship between its inputs, but that also scales to hundreds of thousands of inputs, like ConvNets. The model  leverages an asymmetric attention mechanism to iteratively distill inputs into a tight latent bottleneck, allowing it to scale to handle very large inputs. We show that this architecture is competitive with or outperforms strong, specialized models on classification tasks across various modalities: images, point clouds, audio, video, and video+audio. The Perceiver obtains performance comparable to ResNet-50 and ViT on ImageNet without 2D convolutions by directly attending to 50,000 pixels. It is also competitive in all modalities in AudioSet.
\end{abstract}

\section{Introduction}
\label{submission}

Inductive biases such as spatial locality in early vision are clearly valuable and are famous for drastically increasing the efficiency of learning perceptual models. But, given the increasing availability of large datasets, is the choice to bake such biases into our models with hard architectural decision the correct one? Or are we better off building in as much flexibility as possible, and encouraging the data to speak for itself~\cite{lecun2015deep}?

One glaring issue with strong architectural priors is that they are often modality-specific. For example, if we assume that the input is a single image, we can use our knowledge of its 2D grid structure and build an efficient architecture that relies on 2D convolutional operations. But if we move to a stereo pair, we must decide how to modify this structure to jointly process the pixels from both sensors: should we use an early or late fusion architecture~\cite{karpathy2014largescale} or should we sum or concatenate features? If we move to audio, then the merits of a 2D grid are no longer as clear, and a different type of model, such as 1D convolutions or an LSTM~\cite{hochreiter1997long,6638947}, may be warranted instead. If we want to process point clouds -- a common concern for self-driving cars equippped with Lidar sensors -- then we can longer rely on models that scale best for fixed, low-resolution grids. In short, using standard tools, we are forced to redesign the architecture we use every time the input changes.

\begin{figure*}[t]
    \centering
    \includegraphics[keepaspectratio,width=1\linewidth]{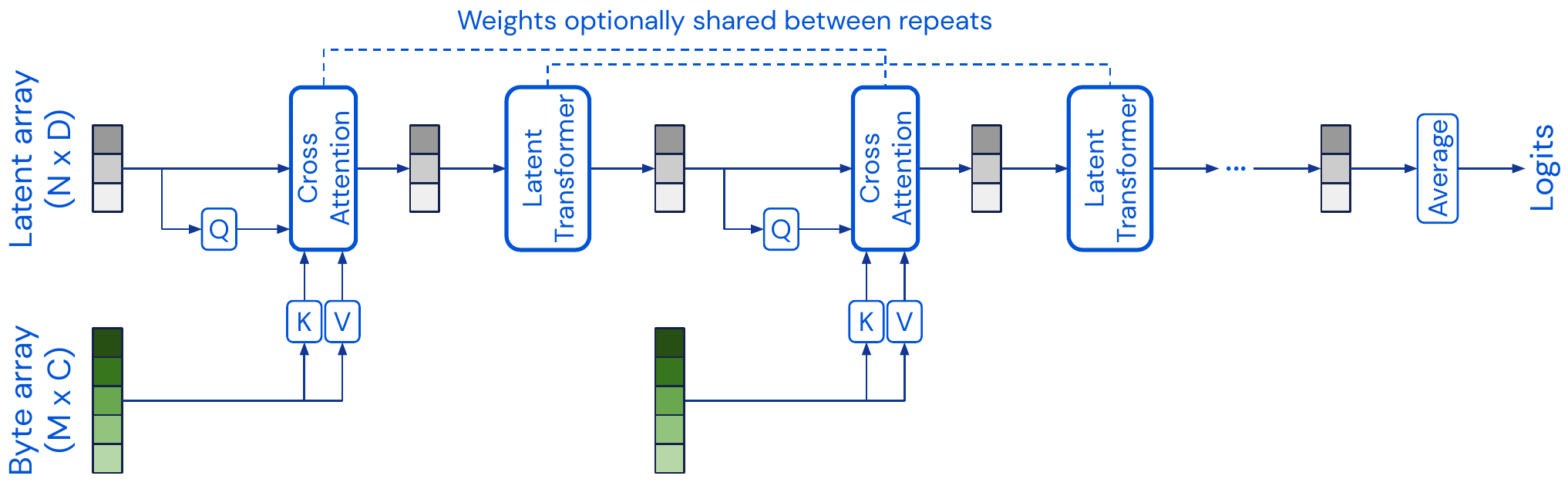}
    \vspace{-10pt}
    \caption{The Perceiver is an architecture based on attentional principles that scales to high-dimensional inputs such as images, videos, audio, point-clouds, and multimodal combinations without making domain-specific assumptions. The Perceiver uses a cross-attention module to project an high-dimensional input byte array to a fixed-dimensional latent bottleneck (the number of input indices $M$ is much larger than the number of latent indices $N$) before processing it using a deep stack of Transformer-style self-attention blocks in the latent space. The Perceiver iteratively attends to the input byte array by alternating cross-attention and latent self-attention blocks.}
    \label{fig:architecture}
    \vspace{-8pt}
\end{figure*}

In this paper we introduce the \textit{Perceiver}, a model designed to handle arbitrary configurations of different modalities using a single Transformer-based architecture. Transformers~\cite{vaswani2017attention} are very flexible architectural blocks that make few assumptions about their inputs, but that also scale quadratically with the number of inputs, in terms of both memory and computation. Recent work has shown impressive performance using Transformers on images, but this work relies on the pixels' grid structure to reduce computational complexity, by first processing pixels using a 2D convolution~\cite{dosovitskiy2020image,touvron2020training}, factorizing the image into columns and rows~\cite{ho2019axial, child2019generating}, or by aggressive subsampling~\cite{chen2020generative}. Instead, we propose a mechanism that can handle high-dimensional inputs while retaining the expressivity and flexibility needed to deal with arbitrary input configurations.

Our core idea is to introduce a small set of latent units that forms an attention bottleneck through which the inputs must pass (Fig.~\ref{fig:architecture}). This eliminates the quadratic scaling problem of all-to-all attention of a classical Transformer and decouples the network depth from the input's size, allowing us to construct very deep models.
By attending to the inputs iteratively, the Perceiver can channel its limited capacity to the most relevant inputs, informed by previous steps. But spatial or temporal information is crucial for many modalities, and it is often essential to distinguish input from one modality or another in multimodal contexts. We can compensate for the lack of explicit structures in our architecture by associating position and modality-specific features with every input element (e.g.\ every pixel, or each audio sample) -- these can be learned or constructed using high-fidelity Fourier features \cite{mildenhall2020nerf, tancik2020fourier, vaswani2017attention}. This is a way of tagging input units with a high-fidelity representation of position and modality, similar to the labeled lined strategy used to construct topographic and cross-sensory maps in biological neural networks by associating the activity of a specific unit with a semantic or spatial location (\citealt{kandel2012principles}, Ch. 21).

We demonstrate performance comparable to strong models such as ResNet-50 and ViT when training on ImageNet for classification; competitive performance on the AudioSet sound event classification benchmark (using raw audio, video, or both); and strong performance relative to comparable approaches on ModelNet-40 point cloud classification. 

\section{Related Work}
\label{sec:related}

\begin{figure*}[t]
\centering
\includegraphics[width=1.0\textwidth]{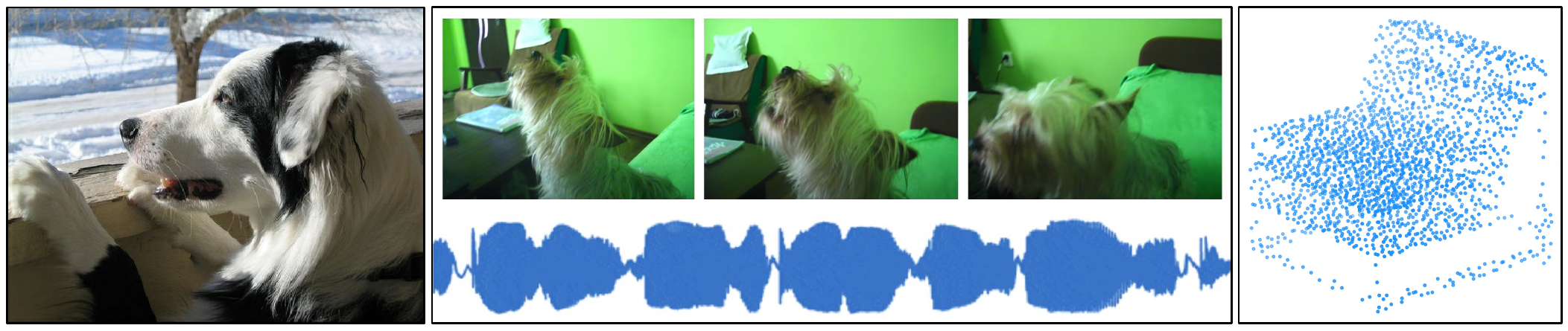}
\vspace{-12pt}
\caption{We train the Perceiver architecture on images from ImageNet~\cite{deng2009imagenet} (left), video and audio from AudioSet~\cite{gemmeke2017audio} (considered both multi- and uni-modally) (center), and 3D point clouds from ModelNet40~\cite{wu20153d} (right). Essentially no architectural changes are required to use the model on a diverse range of input data.}
\label{fig:modalities}
\vspace{-8pt}
\end{figure*}

ConvNets~\cite{fukushima1980selforganizing, lecun1998gradient, ciresan2011highperformance, krizhevsky2012imagenet} have been the dominant family of architectures for perceptual tasks for nearly a full decade, thanks to their good performance and scalability. They can handle high-resolution images while using relatively few parameters and relatively little compute by using convolutions to share weights in 2D and limit each unit's computation to a local 2D neighborhood. However, as discussed in the previous section, they offer limited flexibility when combining multiple signals, unlike the attention-based models dominant in language, as exemplified by Transformers~\cite{vaswani2017attention}.

\noindent \textbf{Efficient attention architectures.}  Transformers are amazingly flexible but scale poorly with the input size because all self-attention layers have the same number of inputs and standard self-attention compares each input to every other input at all layers. Nevertheless, self-attention has been rapidly percolating into perception, for example as pieces of otherwise convolutional models for images~\cite{bello2019attention,cordonnier2019relationship, srinivas2021bottleneck} and videos~\cite{wang2018non,girdhar2019video}. A variety of strategies have been proposed to reduce the size of the input to the Transformer so it can be used on domains that are otherwise too large, including subsampling the input~\cite{chen2020generative} or by first preprocessing the input using convolutions (e.g. \citealt{wu2020visual}). This is the strategy taken by the Vision Transformer (ViT)~\cite{dosovitskiy2020image}, which first reduces the input size to $\sim 200$ using a 2D convolutional layer (referred to as ``linear projection of flattened patches'' in that work) and then applying a Transformer on the resulting inputs along with a BERT-like class token \cite{devlin2018bert}. ViT produces impressive results on ImageNet but this preprocessing strategy restricts it to image-like domains with grid-like sampling patterns. 

Several groups have proposed to modify the internals of the Transformer's self-attention module to gain greater efficiency (see Appendix Sec.~\ref{sec:supp_related} for a discussion). Most closely related to our work is the Set Transformer \cite{lee2019set}. The Set Transformer uses cross-attention to project a large input array to a smaller array, either to reduce the computation within a module or to project inputs to a target output shape (e.g. mapping an input set to logits). Like this work, the Perceiver uses cross-attention over an auxiliary low-dimensional array to reduce the complexity of attention from quadratic to linear in the input size. In a similar vein (but without using cross-attention), Linformer \cite{wang2020linformer} produces linear-complexity self-attention modules by projecting key and value inputs to arrays with a size smaller than the input. Unlike this prior work, the Perceiver uses cross-attention not only to get linear complexity layers, but also to decouple network depth from the input size. As discussed in Sec.~\ref{sec:methods}, it is this decoupling and not merely linear scaling that allows us to build very deep architectures, which appear to be essential for good performance on challenging tasks in a range of domains. We discuss the relationship between the Perceiver and the Set Transformer and related models in more detail in Appendix Sec.~\ref{sec:supp_related}.

\noindent \textbf{Multimodal architectures.}
In current approaches to multimodal processing, separate feature extractors are used for each modality~\cite{kaiser2017one,arandjelovic2018objects,wang2020makes,chen2019uniter,alayrac2020self,lee2020making,xiao2020audiovisual} -- it is generally not sensible to concatenate an audio spectrogram or a raw audio waveform with an image and pass it through a ConvNet. This approach leads to a variety of architectural choices -- such as when to fuse modalities -- that need to be re-tuned for each application. Because of this state of affairs, best-practice architectures for vision cannot be ported to all domains, and specialized models have been developed to handle domains like point clouds~\cite{qi2017pointnet++,guo2020pct}. The Perceiver is designed to very flexibly handle a wide range of inputs out of the box even if they come from very different modalities, including high-bandwidth ones such as images and audio (as illustrated by Fig.~\ref{fig:modalities}).

\section{Methods}
\label{sec:methods}

\subsection{The Perceiver architecture}

\noindent \textbf{Overview.} We build our architecture from two components: (i) a cross-attention module that maps a byte array (e.g.\ an pixel array) and a latent array to a latent array, and (ii) a Transformer tower that maps a latent array to a latent array. The size of the byte array is determined by the input data and is generally large (e.g.\ ImageNet images at resolution 224 have 50,176 pixels), while the size of the latent array is a hyperparameter which is typically much smaller (e.g.\ we use 512 latents on ImageNet). Our model applies the cross-attention module and the Transformer in alternation. This corresponds to projecting the higher-dimensional byte array through a lower-dimension attention bottleneck before processing it with a deep Transformer, and then using the resulting representation to query the input again. The model can also be seen as performing a fully end-to-end clustering of the inputs with latent positions as cluster centres, leveraging a highly asymmetric cross-attention layer. Because we optionally share weights between each instance of the Transformer tower (and between all instances of the cross-attention module but the first), our model can be interpreted as a recurrent neural network (RNN), but unrolled in depth using the same input, rather than in time. All attention modules in the Perceiver are non-causal: we use no masks. The Perceiver architecture is illustrated in Fig.~\ref{fig:architecture}.

\noindent \textbf{Taming quadratic complexity with cross-attention.} We structure our architecture around attention because it is both generally applicable (making less restrictive assumptions about the structure of the input data than e.g.\ ConvNets; it's all you need) and powerful in practice. The main challenge addressed by our architecture's design is scaling attention architectures to very large and generic inputs. Both cross-attention and Transformer modules are structured around the use of query-key-value (QKV) attention \cite{graves2014neural,weston2014memory,bahdanau2015neural}. QKV attention applies three networks -- the query, key, and value networks, which are typically multi-layer perceptrons (MLPs) -- to each element of an input array, producing three arrays that preserve the index dimensionality (or \textit{sequence length}) $M$ of their inputs. The main difficulty of using Transformers on large-scale inputs like images is that the complexity of QKV self-attention is quadratic in the input index dimensionality, but the index dimensionality $M$ of images is typically very large ($M=50176$ for $224 \times 224$ ImageNet images). The challenge is similar for audio: 1 second of audio at standard sampling rates corresponds to around 50,000 raw audio samples. This problem compounds dramatically for multi-modal inputs.

For this reason, prior work that uses attention to process images avoids directly applying standard QKV attention to the input pixel array (see Sec.~\ref{sec:related} and Appendix Sec.~\ref{sec:supp_related} for an overview). Here, we apply attention directly to the inputs by introducing an asymmetry into the attention operation. To see how this works, first note that for $Q \in \mathbb{R}^{M \times D},$ $K \in \mathbb{R}^{M \times C},$ and $V \in \mathbb{R}^{M \times C},$ (where $C$ and $D$ are channel dimensions) the complexity of the QKV attention operation -- essentially, $\text{softmax}(QK^T) V$ -- is $\mathcal{O}(M^2)$, as it involves two matrix multiplications with matrices of large dimension $M$.\footnote{We ignore the contributions of the channel dimensions $C$ and $D$ here, as they are generally small relative to $M$.} So we introduce asymmetry: while $K$ and $V$ are projections of the input byte array, $Q$ is a projection of a learned latent array with index dimension $N \ll M$, where the latent's index dimension $N$ is a hyperparameter. The resulting cross-attention operation has complexity $\mathcal{O}(M N)$.

\noindent \textbf{Uncoupling depth with a latent Transformer.}
The output of the cross-attention module takes the shape of the input to the Q network: that is, the cross-attention layer induces a bottleneck. We exploit this bottleneck by building deep (and hence expressive) Transformers in the latent space: they come at the low cost of $\mathcal{O}(N^2)$. This design allows Perceiver-based architectures to make use of much deeper Transformers than efficient Transformers that use linear-complexity layers, without relying on domain-specific assumptions. This is because a Transformer built on bytes has complexity $\mathcal{O}(LM^2)$ while a latent Transformer has complexity $\mathcal{O}(LN^2)$ (where $N \ll M$), when considered as a function of the number of layers $L$ in addition to index dimensionality. 

This results in an architecture with complexity $\mathcal{O}(MN + LN^2)$, and this is key: by decoupling the input size and the depth, we can add additional Transformer layers at a cost that's independent of the input size. This allows us to construct very large networks on large-scale data. For example, our best ImageNet results use a network with 48 latent Transformer blocks, which is infeasible with networks that couple input size and depth (e.g. see Tab.~\ref{tab:no_transformers}).

Our latent Transformer uses the GPT-2 architecture \cite{radford2019language}, which itself is based on the decoder of the original Transformer architecture \cite{vaswani2017attention}. In our experiments, we use values of $N \le 1024$, which makes our latent Transformer comparable in input size to models in wide-spread use in language. The latent array itself is initialized using a learned position encoding \cite{gehring2017convolutional} (see Appendix Sec.~\ref{sec:supp_arch_details} for details).

\noindent \textbf{Iterative cross-attention \& weight sharing.}
The size of the latent array allows us to directly model pixels and to build deeper Transformers, but the severity of the bottleneck may restrict the network's ability to capture all of the necessary details from the input signal. To hedge against this effect, the Perceiver may be structured with multiple cross-attend layers, which allow the latent array to iteratively extract information from the input image as it is needed. This allows us to tune the model to balance expensive, but informative cross-attends against cheaper, but potentially redundant latent self-attends. As shown in Appendix Tab.~\ref{tab:cross_attend_config}, more cross-attends leads to better performance, but increases the computational requirements of the model because it increases the number of layers with linear dependence on the input size.

Finally, in virtue of the iterative structure of the resulting architecture, we can increase the parameter efficiency of the model by sharing weights between the corresponding blocks of each latent Transformer and/or between cross-attend modules. Latent self-attention blocks can still be shared if only a single cross-attend is used. In our ImageNet experiments, weight sharing results in an approximately 10x reduction in the number of parameters, while reducing overfitting and boosting validation performance. The resulting architecture has the functional form of an RNN with a cross-attention input projection, a bottlenecked latent dimensionality, and a latent Transformer recurrent core. We note that weight sharing has been used for similar goals in Transformers \cite{dehghani2019universal, lan2020albert}.

\subsection{Position encodings}
\label{sec:pos_encodings}

\begin{table}[t]
\centering
\begin{tabular}{|l|l|}
\hline
\textcolor{red}{ResNet-50 \cite{he2016deep}}            & \textcolor{red}{77.6}          \\ 
\textcolor{red}{ViT-B-16 \cite{dosovitskiy2020image}}   & \textcolor{red}{77.9}     \\ \hline
\textcolor{red}{ResNet-50 (FF)}                     & \textcolor{red}{73.5}          \\ 
\textcolor{red}{ViT-B-16 (FF)}                      & \textcolor{red}{76.7}          \\ \hline
\textcolor{blue}{Transformer (64x64, FF)}                   & \textcolor{blue}{57.0}          \\ 
\textcolor{blue}{Perceiver (FF)}                             & \textcolor{blue}{78.0}          \\ \hline
\end{tabular}
\caption{Top-1 validation accuracy (in \%) on ImageNet. \textcolor{red}{Models that use 2D convolutions} exploit domain-specific grid structure architecturally, while \textcolor{blue}{models that only use global attention} do not. The first block reports standard performance from pixels -- these numbers are taken from the literature. The second block shows performance when the inputs are RGB values concatenated with 2D Fourier features (FF) -- the same that the Perceiver receives. This block uses our implementation of the baselines. The Perceiver is competitive with standard baselines on ImageNet without relying on domain-specific architectural assumptions.}
\label{tab:imagenet}
\vspace{-12pt}
\end{table}

\begin{table}[t]
\centering
\begin{tabular}{|l|l|l||l|}
\hline
                                      & Raw           & Perm.         & Input RF   \\ \hline
\textcolor{red}{ResNet-50 (FF)}       & \textcolor{red}{73.5}          & \textcolor{red}{39.4}          &  \textcolor{red}{49}  \\ 
\textcolor{red}{ViT-B-16 (FF)}                         & \textcolor{red}{76.7}          & \textcolor{red}{61.7}          &  \textcolor{red}{256}  \\  \hline
\textcolor{blue}{Transformer (64x64) (FF)}               & \textcolor{blue}{57.0}          & \textcolor{blue}{57.0}          &  \textcolor{blue}{4,096}  \\
\textcolor{blue}{Perceiver}:                            &               &               & \\
\hspace{1em} \textcolor{blue}{(FF)}                      & \textcolor{blue}{78.0}          & \textcolor{blue}{78.0}          & \textcolor{blue}{50,176} \\
\hspace{1em} \textcolor{blue}{(Learned pos.)}            & \textcolor{blue}{70.9}          & \textcolor{blue}{70.9}          & \textcolor{blue}{50,176} \\ \hline
\end{tabular}
\vspace{-8pt}
\caption{Top-1 validation accuracy (in \%) on standard (raw) and \textbf{permuted} ImageNet (higher is better). Position encodings (in parentheses) are constructed before permutation, see text for details. While \textcolor{blue}{models that only use global attention} are stable under permutation, \textcolor{red}{models that use 2D convolutions} to process local neighborhoods are not. The size of the local neighborhood at input is given by the input receptive field (RF) size, in pixels.}
\label{tab:imagenet_permuted}
\vspace{-15pt}
\end{table}

\noindent \textbf{Permutation invariance and position information.} Attention is a permutation-invariant operation, and this property is preserved by the Perceiver and related models \cite{lee2019set}. A pure attention model will return the same output regardless of the order of its inputs, leaving no trace of the input's ordering on its outputs. This property makes attention-based architectures well-suited for many types of data, as they make no assumptions about which spatial relationships or symmetries to prioritize. In contrast, the ConvNets that are typically used in image processing -- such as residual networks (ResNets) \cite{he2016deep} -- bake in 2D spatial structure in several ways, including by using filters that look only at local regions of space (which makes it easier to capture the relationship between nearby pixels than between distant pixels), by sharing weights across both spatial dimensions (which helps to model data with statistics that are invariant to translation), and by repeatedly applying small filters (which helps to model data with statistics that are invariant to scale).

But permutation invariance means that the Perceiver's architecture cannot in and of itself exploit spatial relationships in the input data. Spatial relationships are essential for sensory reasoning \cite{kant1781critique} and this limitation is clearly unsatisfying. In the attention literature, position information is typically injected by tagging \textit{position encodings} onto the input features \cite{vaswani2017attention}; we pursue this strategy here as well. While position information is typically used to encode sequence position in the context of language, it can also be used to encode spatial, temporal, and modality identity.

\noindent \textbf{Scalable Fourier features.} Here, we use a strategy that has recently gained renewed prominence, both in language and in vision: Fourier feature position encodings \cite{stanley2007compositional, vaswani2017attention, parmar2018image, tancik2020fourier, mildenhall2020nerf}. We use a parameterization of Fourier features that allows us to (i) directly represent the position structure of the input data (preserving 1D temporal or 2D spatial structure for audio or images, respectively, or 3D spatiotemporal structure for videos), (ii) control the number of frequency bands in our position encoding independently of the cutoff frequency, and (iii) uniformly sample all frequencies up to a target resolution. 

We parametrize the frequency encoding to take the values $[\sin(f_k \pi x_d), \cos(f_k \pi x_d)]$, where the frequency $f_k$ is the $k^{\text{th}}$ band of a bank of frequencies spaced equally between 1 and $\frac{\mu}{2}$. $\frac{\mu}{2}$ can be naturally interpreted as the Nyquist frequency \citep{nyquist1928certain} corresponding to a target sampling rate of $\mu$. By allowing the network to resolve all positions in an input array, we can encourage it to learn to compare the values of bytes at any positions in the input array. $x_d$ is the value of the input position along the $d^{\text{th}}$ dimension (e.g.\ for images $d=2$ and for video $d=3$). $x_d$ takes values in $[-1, 1]$ for each dimension. We concatenate the raw position value $x_d$ to produce the final representation of position. This results in a position encoding of size $d(2K + 1)$.

This parameterization is related to the NeRF position encoding scheme \cite{mildenhall2020nerf}, which is built around frequency bands with increasing powers of two (the $k^{\text{th}}$ band has frequency $2^k$). This leads to very high frequencies for even modest numbers of bands, and in some experiments, we encountered numerical instability when using this parameterization beyond around $k=15$ bands. 

In language modelling, Transformer inputs are typically produced by adding a position encoding to the input encoding (the size of the position encoding is tailored to the encoding used). We found it beneficial to instead concatenate the position and input features before passing them into the Perceiver. This difference is perhaps explained by the fact that input features in language tend to be larger and sparser than the modalities considered here.

\noindent \textbf{Position encodings are generally applicable.} Does the use of position encodings undermine our claim to be moving from a more domain-specific architecture built to exploit 2D structure to a more general ones? No, for three reasons. \textbf{First,} while the architectural imposition of position information hard codes a specific positional prior, the feature-based approach allows the network to learn how to use (or ignore) the position structure. This is in accord with the idea that greater generality follows from making as much of a system learnable as possible \cite{sutton2019bitter}. \textbf{Second,} it is possible to redesign architectural priors for data domains with different structures, such as videos \cite{tran2015learning} or audio~\cite{ford2019deep}, or for groups other than the group of linear translations (e.g. \citealt{cohen2016group, bronstein2017geometric, esteves2018learning})); this however often requires a tremendous amount of researcher time and expertise. In contrast, a position encoding can be easily adapted to a new domain: Fourier features are trivial to adapt as long as the input dimensionality is relatively small and known. In the broader Transformer literature, simple learned position encoding have proven to be sufficient for good results in many settings. We find that a similar strategy produces reasonable results on ImageNet (see Table~\ref{tab:imagenet_permuted}, bottom row) even though it has no knowledge whatsoever about the input 2D structure. \textbf{Third,} position encodings can be naturally extended to multimodal data: each domain can use a position encoding with the correct dimensionality for its data, with learned encodings used to distinguish domains (we use this strategy for multimodal Audioset, see Sec.~\ref{sec:audioset}).

\section{Experiments}

The next few subsections are organized by the modalit\{y, ies\} used (illustrated in Fig.~\ref{fig:modalities}). We evaluate the effect of model configuration and hyperparameters on ImageNet classification in the supplement (Sec.~\ref{sec:supp_ablations}). As baselines we consider ResNet-50~\cite{he2016deep}, a very widely model for both vision and audio and possibly the closest thing to a general perceptual architecture so far. We also consider two Transformer variants, the recently proposed ViT~\cite{dosovitskiy2020image}, and a stack of Transformers~\cite{vaswani2017attention}. All experiments were conducted using JAX \cite{jax2018github} and the DeepMind JAX ecosystem \cite{deepmind2020jax}.

\subsection{Images -- ImageNet}

\begin{figure*}
    \centering
    \includegraphics[keepaspectratio,width=0.9\linewidth]{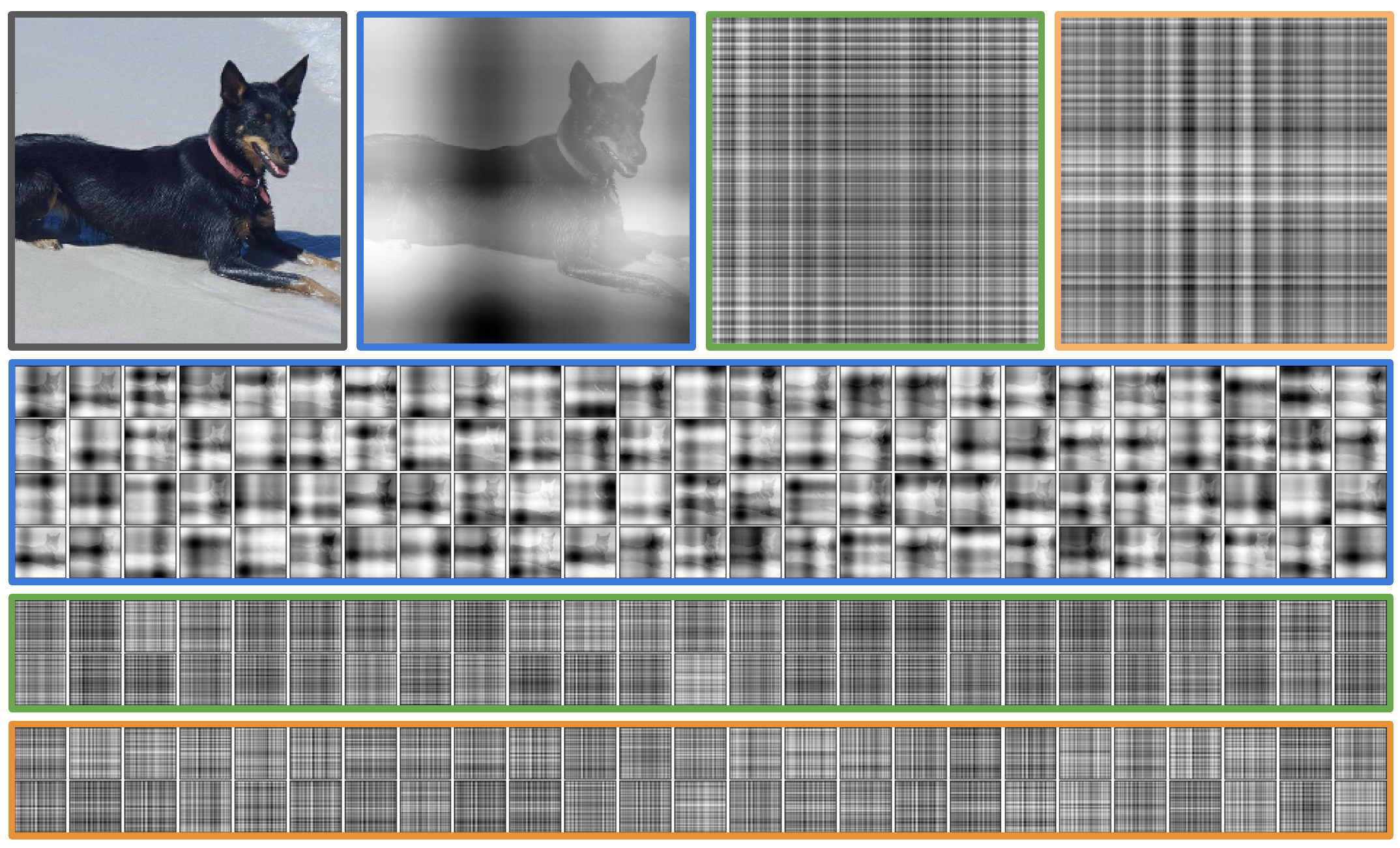}
    \vspace{-12pt}
    \caption{Attention maps from the \textbf{\textcolor{my_blue}{first}}, \textbf{\textcolor{my_green}{second}}, and \textbf{\textcolor{orange}{eighth}} (final) cross-attention layers of a model on ImageNet with 8 cross-attention modules. Cross-attention modules 2-8 share weights in this model. \textbf{Row 1:} Original image and close-ups of one attention map from each of these layers. \textbf{Rows 2-4:} Overview of the attention maps of the cross-attention modules. Attention maps appear to scan the input image using tartan-like patterns at a range of spatial frequencies. The visualized attention maps are \textit{not} overlaid on the input image: any apparent image structure is present in the attention map itself (the dog is clearly visible in several of the first module's attention maps).}
    \label{fig:attention_map}
    \vspace{-12pt}
\end{figure*}

First, we consider the task of single-image classification using the ILSVRC 2012 split of the ImageNet dataset \cite{deng2009imagenet}. ImageNet has been a crucial bellwether in the development of architectures for image recognition~\cite{krizhevsky2012imagenet,simonyan2014very,szegedy2015going,he2016deep} and, until recently, it has been dominated by ConvNet architectures. Each image on ImageNet has a single label so we use softmax outputs and a cross-entropy loss to train for the classification task. As is standard practice, we evaluate our model and all baselines using the top-1 accuracy on the held-out validation set (the test set is not publicly available). We train our model using images sampled by Inception-style preprocessing \cite{szegedy2015going}, including standard $224 \times 224$ pixel crops. Additionally, we augment all images using RandAugment \cite{cubuk2020randaugment} at training time.

\noindent \textbf{Position encodings.} We generate position encodings by first using the (x, y) positions on the $224 \times 224$ input crop. (x, y) coordinates are standardized to [-1, 1] for each dimension of a crop (see Appendix Fig.~\ref{fig:cropping}). In Inception-style preprocessing, the raw crop can have a non-uniform aspect ratio, which may lead to aspect ratio distortion in both the input crop and in the (x, y) coordinates used to generate the position encoding. In early experiments, we tried using image coordinates rather than crop coordinates as the basis of the position encoding, but we found that this led to model overfitting. We suspect that this occurs because the Perceiver's architecture may allow it to memorize training-set image by latching onto a small number of input pixels, if they are always associated with the same (RGB, position) feature. By using crops, we effectively introduce augmentation in both position and aspect ratio, which breaks correlations between RGB values and position features and makes it much harder to associate an image label with a small number of pixels.

\noindent \textbf{Optimization and hyperparameters.} Although it is typical to train convolutional networks on ImageNet using SGD, we found it easier to optimize Perceiver models using the LAMB optimizer \cite{you2020large}, which was developed for optimizing Transformer-based models. We trained models for 120 epochs with an initial learning rate of 0.004, decaying it by a factor of 10 at [84, 102, 114] epochs. The best-performing Perceiver we identified on ImageNet attends to the input image 8 times, each time processing the full 50,176-pixel input array using a cross-attend module and a latent Transformer with 6 blocks and one cross-attend module with a single head per block. We found that sharing the initial cross-attention with subsequent cross-attends led to instability in training, so we share all cross-attends after the first.  The dense subblock of each Transformer block doesn't use a bottleneck. We used a latent array with 512 indices and 1024 channels, and position encodings generated with 64 bands and a maximum resolution of 224 pixels. On ImageNet, we found that models of this size overfit without weight sharing, so we use a model that shares weights for all but the first cross-attend and latent Transformer modules. The resulting model has $\sim 45$ million parameters, making it comparable in size to convolutional models used on ImageNet.

\noindent \textbf{Standard ImageNet.} As shown in Table~\ref{tab:imagenet}, the Perceiver model we trained on ImageNet obtains results that are competitive with models specifically designed for processing images. We include ResNet-50 results from \cite{cubuk2020randaugment}, as these numbers use RandAugment and hence better match our training protocol. To account for the Perceiver's use of Fourier features at input, we trained versions of the benchmark models with this input as well and found that it produced comparable, if slightly worse, performance to models trained solely on RGB input. Additionally, we tested the performance of a pure Transformer model. Because Transformers cannot handle ImageNet-scale data, we first downsampled the input images to $64 \times 64$ before passing it into the Transformer (we obtained similar results using 96x96 inputs, which however is much slower to train and more memory-intensive so we could not use as many layers). The Transformer model we consider has the same architecture as the latent Transformer of the Perceiver, differing only in hyperparameters (we swept each model independently), for more details please consult the Appendix. Note that state-of-the art on ImageNet without pretraining was 86.5\% top-1 accuracy at submission \cite{brock2021high}.

\noindent \textbf{Permuted ImageNet.} To evaluate how important domain-specific assumptions about grid structure are to the performance of the benchmark methods, we evaluate all methods on permuted ImageNet. To generate permutations, we use a single, shared permutation pattern for all images. The permutation is performed \textit{after} position features are generated. We make this choice because it still allows each network to infer the spatial relationship between points (using the position encoding), but prevents the network from using an architectural inductive bias to do so. In other words, under these conditions, networks that use 2D convolutions cannot exploit the local neighorhood structure of inputs to reason about space, but must reason in terms of features, just like structure-agnostic architectures like Transformers and Perceivers. The results of this experiment are shown in Table~\ref{tab:imagenet_permuted}. As the Transformer and Perceiver effectively treat any input as a permuted input, their results are not affected, but we find that the performance of both ViT and ResNet suffer dramatically after permutation, even though these models have access to position encodings. Both of these models still perform far above chance, suggesting they are able to reason in terms of position features. ViT is more stable under permutation than the ResNet model: this is likely because ViT uses a single 2D convolution layer with a fairly large receptive field ($16 \times 16 = 256$ pixels) followed by a Transformer, while ResNet-50 uses an initial convolution with a relatively small receptive field ($7 \times 7 = 49$ pixels) and an architecture essentially entirely composed of convolutions.

The Perceiver architecture itself makes no assumptions about the spatial structure of its input, but the Fourier features position encoding we use by default does. By replacing these features with a fully learned, 128-dimensional position encoding, we can evaluate the performance of a Perceiver \textbf{with no knowledge of the spatial structure of the inputs}. The results of this experiment are shown in the bottom row of Table~\ref{tab:imagenet_permuted}. The position encoding used here is initialized randomly and trained end-to-end along with the network (using the same initialization type used for the latent array's position encoding, see Appendix Sec.~\ref{sec:supp_arch_details}). Because the position encodings used here are unaware of the structure of the input, it makes no difference whether inputs are permuted before or after the position encoding is constructed. We found that the network with 8 cross-attends had stability issues when learned position encodings are used, so we report results from a network with a single cross-attend.

\begin{table*}[]
\centering
\begin{tabular}{|l|l|l|l|l|}
\hline
Model / Inputs                                      & Audio        & Video         & A+V            \\ \hline
Benchmark~\cite{gemmeke2017audio}                   & 31.4         &  -            &  -             \\ 
Attention~\cite{kong2018audio}                      & 32.7         & -             & -              \\ 
Multi-level Attention ~\cite{yu2018multi}           & 36.0         & -             & -              \\ 
ResNet-50~\cite{ford2019deep}                       & 38.0         &  -            &  -             \\ 
CNN-14~\cite{kong2020panns}                         & 43.1         &  -            &  -             \\
CNN-14 (no balancing \& no mixup) ~\cite{kong2020panns} & 37.5       &  -            &  -             \\ \hline
G-blend~\cite{wang2020makes}                        & 32.4         & 18.8          & 41.8           \\
Attention AV-fusion~\cite{fayek2020large}           & 38.4         & 25.7          & 46.2           \\ \hline
Perceiver (raw audio) & 38.3 & 25.8 & 43.5 \\
Perceiver (mel spectrogram) & 38.4 & 25.8 & 43.2 \\ 
Perceiver (mel spectrogram - tuned) & - & - & 44.2 \\ \hline
\end{tabular}
\caption{Perceiver performance on AudioSet, compared to state-of-the-art models (mAP, higher is better).}
\label{tbl:audioset}
\vspace{-8pt}
\end{table*}

On the face of it, this experiment may appear contrived -- we know the grid structure, so why don't we use it? But the permuted settings provides a convenient model of the challenges presented by modalities that are challenging and large-scale (like ImageNet) but aren't naturally mapped to a 2D grid (e.g. point clouds, olfactory inputs, touch inputs, Lidar, etc.) or that include modalities that don't share a common grid (e.g. images + language, video + audio, somatosensory inputs + motor feedback, etc.).

\noindent \textbf{Attention maps.} Fig.~\ref{fig:attention_map} visualizes the attention maps at several cross-attention modules for a sample image from ImageNet (we include additional results in the Appendix). Each attention map shows the output of the $QK^T$ operation at each of the model's 512 latent indices and each input pixel. We visualize attention maps before the softmax, as the softmax outputs are very sparse and hard to interpret. This model uses unshared weights in its initial cross-attention, but shares weights for all subsequent layers. The initial and later cross-attention layers produce qualitatively different attention maps: while the early modules shows clear traces of the input image (the dog pops out in many attention maps), the attention maps of later modules manifest as high-frequency plaid lattices. While the attention maps for modules 2 and 7 show similar structure, the specific details of corresponding maps do vary, which suggests the network attends to different sets of pixels at subsequent stages. The banded, variable-frequency structure of the attention maps appears to reflect the spatial frequency structure of the Fourier feature position encodings used on ImageNet. This tartan-like pattern is not present in networks with fully learned position encodings, suggesting it is at least in part due to the Fourier features. 

\subsection{Audio and video -- AudioSet}
\label{sec:audioset}

We experimented with audio event classification in video using AudioSet~\cite{gemmeke2017audio}, a large dataset with 1.7M 10s long training videos and 527 classes. Videos may have multiple labels so we use a sigmoid cross entropy loss and evaluate using mean average precision (mAP). We evaluate the Perceiver using audio (using either the raw audio waveform or mel spectrogram), video, and audio + video as inputs. We sample 32-frame clips (1.28s at 25fps) in training; for evaluation we split the videos into 16 overlapping 32-frame clips, covering the whole 10s, and average the scores. We train models for 100 epochs.

Given the scale of the dataset we used a faster version of the ImageNet model with only 2 attention iterations instead of 8, but 8 self-attention layers per Transformer block instead of 6. We omit weight sharing to compensate for the smaller size. We experimented briefly with temporal unrolling -- e.g. processing one frame per cross-attend -- and found that it worked well and efficiently for video, but hurt performance for audio. Audio may require longer attention context.

\noindent \textbf{Audio only.} We use audio sampled at 48Khz resulting in 61,440 audio samples over 1.28s of video. We experimented with two settings: in the first we divide the raw signal into segments of 128 elements, for a total of 480 128-d vectors and input these to the Perceiver; the second setting uses a mel spectrogram resulting in 4800 inputs to the Perceiver, once flattened. As augmentations, for raw audio we simply sample in time, consistently with the video sampling. For spectrograms we use also  specaugment~\cite{park2019specaugment}.

\noindent \textbf{Video.} A full 32 frame clip at 224x224 resolution has more than 2 million pixels.  We experimented using tiny space-time patches with dimensions 2x8x8, resulting in a total of 12,544 inputs to the Perceiver. 
We compute Fourier features for horizontal, vertical and time coordinates (scaled to [-1, 1]), and concatenated them with the RGB values. We use the same model as in the audio experiments but now taking space-time patches as input rather than audio. We performed color augmentation, inception-type resizing, randomly flipping, and cropped to 224x224 resolution.

\noindent \textbf{Audio + video.} In this experiment we feed the Perceiver both the 12,544 space-time patches and either 480 raw audio vectors or 4,800 spectrogram values. Since modalities are fused at input, audio and video inputs need to have the same number of channels. We achieve this by concatenating a learned, modality-specific encoding to each input. As video has more channels, we use an embedding of size 4 for video inputs and make the audio encoding as large as necessary for the input channels between the two input arrays. This encoding doubles as a modality-specific position encoding (as discussed in Sec.~\ref{sec:pos_encodings}), and we found it worked better than simply passing the audio encoding through a linear layer to match the video. Another thing that proved useful was \textbf{video dropout} -- entirely zeroing out the video stream during training with some probability -- a 30\% probability for each example in each batch worked well. This may help the network to not overfit to video: these inputs provide a larger but less discriminative signal on Audioset. We observed a more than 3\% improvement by using this procedure; without it the spectrogram-based model scored 39.9\% mAP (vs. 43.2\%) and the raw audio model scored 39.7\% (vs. 43.5\%). After the ICML camera-ready deadline we tuned the spectrogram model further, and improved results to 44.2 by turning off specaugment and also dropping the spectrogram modality with 10\% probability.

\noindent \textbf{Results.} Table~\ref{tbl:audioset} shows that the Perceiver obtains near state-of-the-art results on both video- and audio-only experiments. On raw audio the Perceiver gets 38.4, better than most ConvNet models except CNN-14 ~\cite{kong2020panns}  which uses extra   AugMix~\cite{hendrycks2019augmix} and class-balances the data -- we hope to incorporate this in future work. Without these improvements the CNN-14 model does slightly worse than the Perceiver (37.5 mAP). Most previous methods use spectrograms as input but we find we can obtain similar performance even when using raw audio.

Audio+video fusion leads to solid improvements over single modalities (and outperforms specialized fusion optimization approaches~\cite{wang2020makes}) but is still lower than the state-of-the-art approach that uses separate models with late fusion~\cite{fayek2020large}. We will investigate this in future work. We visualize video and audio attention maps in Appendix Sec.~\ref{sec:audiovisual_viz}.

\subsection{Point clouds -- ModelNet40}

ModelNet40~\cite{wu20153d} is a dataset of point clouds derived from 3D triangular meshes spanning 40 object categories. The task is to predict the class of each object, given the coordinates of $\sim$ 2000 points in 3D space. ModelNet is small compared to other datasets used in our experiments: it has 9,843 training examples and 2,468 testing examples. To apply our model, we first preprocess point clouds by zero-centering them. To augment in training we apply random per-point scaling (between 0.99 and 1.01) followed by zero-mean and unit-cube normalization. We also explored random per-point translation (between -0.02 and 0.02) and random point-cloud rotation, but we found this did not improve performance.

\begin{table}[t]
\centering
\begin{tabular}{|l|l|l|l|}
\hline
                                                      & Accuracy                         \\ \hline
\textcolor{red}{PointNet++~\cite{qi2017pointnet++}}   &  \textcolor{red}{\textbf{91.9}}  \\ \hline
\textcolor{blue}{ResNet-50 (FF)}                      &  \textcolor{blue}{66.3}          \\        
\textcolor{blue}{ViT-B-2 (FF)}                        &  \textcolor{blue}{78.9}          \\        
\textcolor{blue}{ViT-B-4 (FF)}                        &  \textcolor{blue}{73.4}          \\        
\textcolor{blue}{ViT-B-8 (FF)}                        &  \textcolor{blue}{65.3}          \\        
\textcolor{blue}{ViT-B-16 (FF)}                       &  \textcolor{blue}{59.6}          \\        
\textcolor{blue}{Transformer (44x44)}                 &  \textcolor{blue}{82.1}          \\ \hline
\textcolor{blue}{Perceiver}                           &  \textcolor{blue}{\textbf{85.7}} \\ \hline

\end{tabular}
\caption{Top-1 test-set classification accuracy (in \%) on ModelNet40. Higher is better. We report best result per model class, selected by test-set score. There are no RGB features nor a natural grid structure on this dataset. We compare to the generic baselines considered in previous sections with Fourier feature encodings of positions, as well as to a specialized model: PointNet++~\cite{qi2017pointnet++}. \textcolor{red}{PointNet++} uses extra geometric features and performs more advanced augmentations that we did not consider here and are not used for the models in \textcolor{blue}{blue}. 
\label{modelnet40}}
\vspace{-12pt}
\end{table}

We used an architecture with 2 cross-attentions and 6 self-attention layers for each block and otherwise used the same architectural settings as ImageNet.  We used a higher maximum frequency than for image data to account for the irregular sampling structure of point clouds - we used a max frequency of 1120 ($10\times$ the value used on ImageNet). We obtained the best results using 64 frequency bands, and we noticed that values higher than 256 generally led to more severe overfitting. We used a batch size of 512 and trained with LAMB with a constant learning rate of \num{1e-3}: models saturated in performance within 50,000 training steps.

Note that state-of-the-art methods on this benchmark are quite small and specialized and typically perform much more sophisticated data augmentation / feature engineering procedures, including fitting surfaces to the point clouds and using face normals as additional points~\cite{qi2017pointnet++}. Here we are mostly interested in comparing to more generic models such as the ImageNet baselines and to assess how the various models deal with data that does not conform to a grid. Results of the Perceiver compared to the baselines are shown in Tab.~\ref{modelnet40}. We arrange each point cloud into a 2D grid randomly, then feed it through each model. For ViT we varied the size of the patch size used at input.

\section{Discussion}

We have presented the Perceiver, a Transformer-based model that scales to more than a hundred thousand inputs. This opens new avenues for general perception architectures that make few assumptions about their inputs and that can handle arbitrary sensor configurations, while enabling fusion of information at all levels.

With great flexibility comes great overfitting, and many of our design decisions were made to mitigate this. In future work, we would like to pre-train our image classification model on very large scale data~\cite{dosovitskiy2020image}. We obtain strong results on  the large AudioSet dataset, which has 1.7M examples and where the Perceiver performed competitively with strong and recent state-of-the-art entries on audio, video and both combined. On ImageNet the model performs on par with ResNet-50 and ViT. When comparing these models across all different modalities and combinations considered in the paper, the Perceiver does best overall.

While we reduced the amount of modality-specific prior knowledge in the model, we still employ modality-specific augmentation and position encoding. End-to-end modality-agnostic learning remains an interesting research direction. 

\section*{Acknowledgements}

We are grateful to Sander Dieleman and Matt Botvinick for reviewing drafts of the paper, to Adri\`{a} Recasens Continente and Luyu Wang for help with AudioSet (especially Luyu for identifying an evaluation bug we had), and to Chris Burgess, Fede Carnevale, Mateusz Malinowski, Lo\"{i}c Matthey, David Pfau, Adam Santoro, Evan Shelhamer, Greg Wayne, Chen Yan, Daniel Zoran and others at DeepMind for helpful conversations and suggestions. We thank Irwan Bello, James Betker, Andreas Kirsch, Christian Szegedy, Weidi Xie and others for comments on an earlier draft.

\bibliography{perceiver}
\bibliographystyle{icml2021}

\clearpage

\appendix
\appendixpage

\section{Extended related work}
\label{sec:supp_related}

\noindent \textbf{Efficient attention architectures, cont'd.}
Several strategies have been proposed to gain greater efficiency by modifying the internals of the Transformer's self-attention module, including using local or patchwise self-attention \cite{parmar2018image, ramachandran2019standalone, zhao2020exploring, sukhbaatar2019adaptive}, using non-local, non-dense attention patterns~\cite{ho2019axial,wang2020axial, beltagy2020longformer, child2019generating, correia2019adaptively, ye2019bptransformer, roy2020efficient}, approximating or otherwise simplifying the matrix multiplications used in QKV attention \cite{choromanski2021rethinking, peng2021random, kitaev2020reformer, xiong2021nystromformer, katharopoulos2020transformers, tay2021synthesizer}, or by introducing bottlenecks into each module's computation \cite{lee2019set, wang2020linformer}. The primary contribution of this body of work is a set of modules with similar flexibility to the Transformer's densely-connected self-attention block, but at sub-quadratic computational cost (see \citealt{tay2020efficient, tay2020long} for more detailed reviews). The focus of our work is primarily on producing an architecture that is efficient as a whole and is suitable for many domains, rather than improving the complexity of the Transformer's self-attention module itself. In this sense, our work is complementary to this large and very interesting body of work, and it is likely that some of these approaches could be used to further increase the Perceiver's efficiency.

\noindent \textbf{Relationship to the Set Transformer.}
The Set Transformer work \cite{lee2019set} introduces two modules (called ISAB for ``induced set attention block'' and PMA for ``pooling by multiheaded attention''), that function similarly to the cross-attention blocks we use here but are deployed in a different manner. ISAB is used to map an input array (interpreted as a set) to a low-dimensional array and immediately map it back to the input space. Stacking these blocks leads to an architecture that scales linearly in compute/memory with input size like the Perceiver's cross-attention module, but without the advantage of the Perceiver's latent array (which completely decouples the cost of the latent Transformer from the input size): a fully-ISAB model scales as $\mathcal{O}(LMN)$, rather than $\mathcal{O}(MN + LN^2)$, like the Perceiver (where $M$ is the index dimension of the input, $N$ the index dimension of the latent, and $L$ the network depth). 

PMA is used to map an input array to an output array with a sized determined by the task (e.g. 1 point for classification or 4 points for a 4-way clustering task). It is used to map to a target output size and not to induce a latent space. In contrast, the Perceiver's latent space has a size that is independent of the task (it is a hyperparameter, and typically much larger than the task output) and is designed specifically to facilitate the efficient construction of deep, latent Transformers. To use the Set Transformer terminology, a Perceiver directly feeds its input to a PMA-like block (or $\frac{1}{2}$ of an ISAB-like block) whose output size is relatively large (e.g. 512) and task-independent rather than determined by the task; it would be 1 (for classification) if used as proposed in the Set Transformer. This is followed by a very deep stack of (latent) self-attention blocks and a final average and project. In other words, Perceivers exploit similar primitives to the Set Transformer, but compose them differently, in service of building an architecture with improved scaling properties.

\noindent \textbf{Cross-attention and attentional latents.}
More broadly, cross-attention has been used to augment Transformer architectures with attention to the longer-horizon past \cite{rae2020compressive, dai2019transformerxl} and to produce architectures that write to and/or read from fixed-size arrays or memories \cite{santoro2018relational, goyal2021coordination}, all while keeping the cost of each operation linear in the input size. We use cross-attention to induce a latent space for deep processing. This can be viewed as a fully attentional, domain-agnostic analogue of models that stacks self-attention on top of convolutional feature maps to perform cheap but global processing on top or in conjunction with otherwise spatially localized convolutional feature maps (e.g. \citealt{carion2020detr, locatello2020object, wang2021maxdeeplab}).

\noindent \textbf{Global, re-entrant processing.} The Perceiver performs global computations from the first layer: although contemporary architectures typically first process locally, the notion of building perception systems using global processing has a long history (e.g. \citealt{kohler1967gestalt,shi2000normalized}). When inputs grow very large, this may introduce a bandwidth bottleneck. By using multiple cross-attentions, the Perceiver can use a form of re-entrant processing to mitigate this effect, by allowing first-pass processing of an input to feed back and influence how the input is processed in subsequent passes. Re-entrant processing of this kind (sometimes referred to as top-down processing) has a long history in computer vision~\cite{borenstein2004combining,kumar2005obj,carreira2016human,hu2016bottom,yang2018convolutional,lin2020context}. There is widespread evidence that it plays an important role in human vision (e.g. \citealt{felleman1991distributed, olshausen1993neurobiological, lollo2000competition}), which is characterized by limited bandwidth input streams \cite{wolfe2006changes}. In the Perceiver, attention to the full set of inputs can be influenced by a latent array produced by previous iterations of the model, allowing the model focus on subsets of inputs that are most promising in a soft way~\cite{zoran2020towards}.

\section{Ablations}
\label{sec:supp_ablations}

\begin{figure}[]
    \centering
    \subfigure[Crop-relative coordinates]{\label{fig:a}\includegraphics[width=0.23\textwidth]{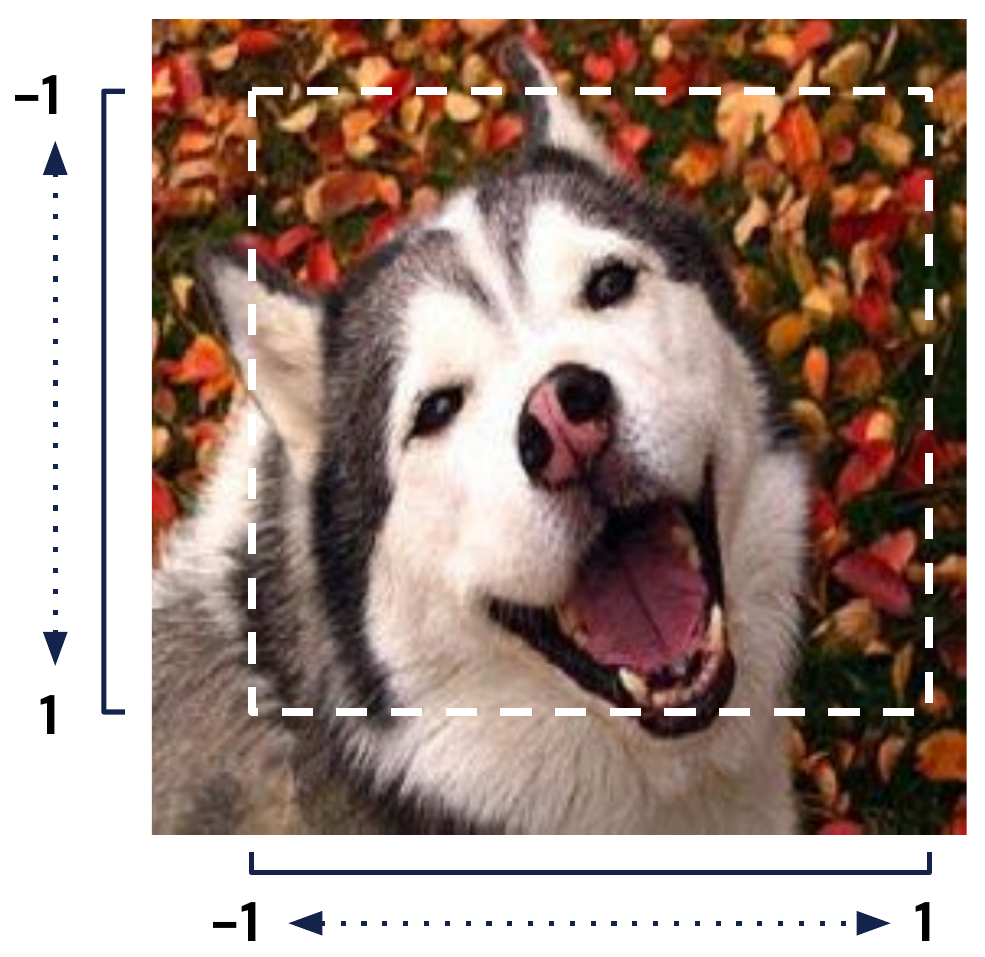}}
    \subfigure[Image-relative coordinates]{\label{fig:b}\includegraphics[width=0.23\textwidth]{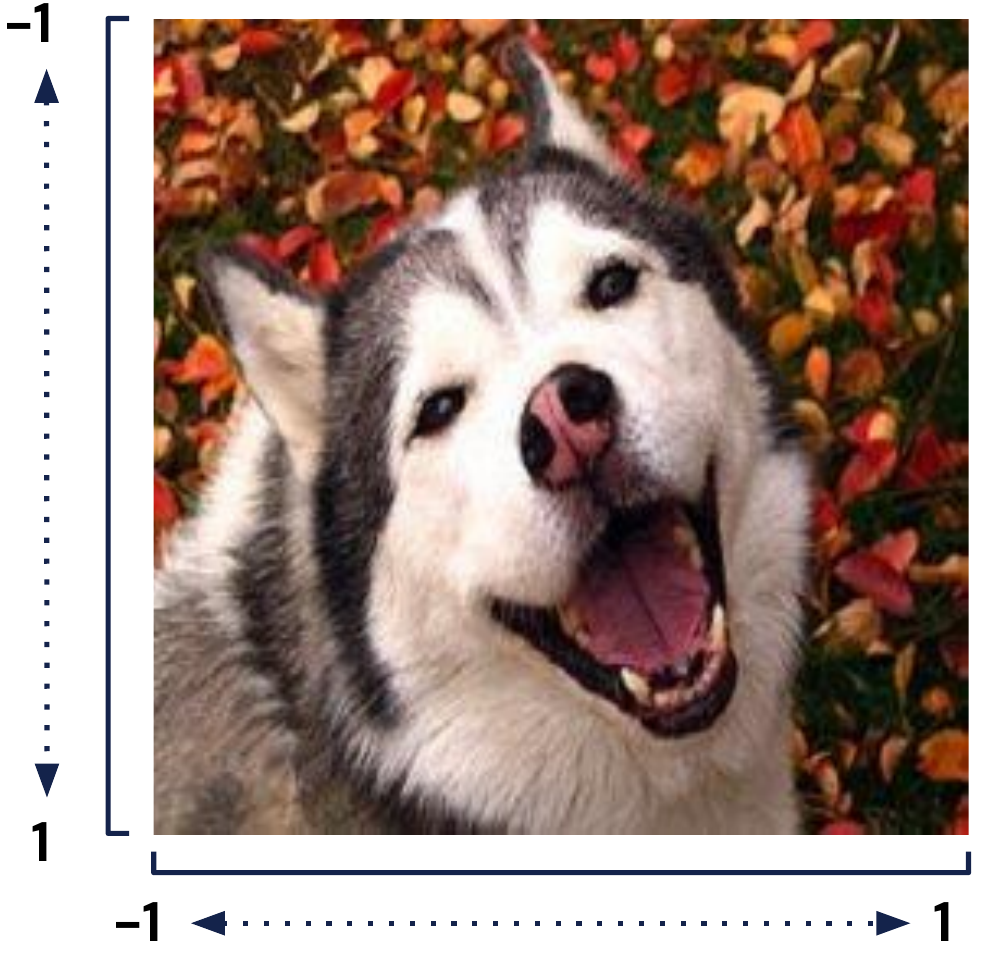}}
    \caption{For ImageNet experiments, we generate position encodings using [-1, 1]-normalized (x, y)-coordinates drawn from (a) crops rather than from (b) the raw images, as we find the latter leads to overfitting.}
    \label{fig:cropping}
    \vspace{-6pt}
\end{figure}

To illustrate the effect of various network hyperparameters, we considered a small Perceiver model and swept a number of options around it. Unlike ConvNets, each module in a Perceiver-based architecture takes as input the full input byte array. This makes it possible to sweep processing hyperparameters (e.g.\ depth, capacity, etc.), without reducing the effective receptive field size of the network as a whole. The base model did not share either self-attention or cross-attention parameters, used 8 heads per self-attention module, 4 self-attention modules per block, performed 2 cross-attends per image, and had a latent with an index dimension of 512 and a channel dimension 512. We used a small batch size of 64 across 32 TPUs to make sure all models fit comfortably in memory no matter how extreme the parameters. We trained all models for 5 million steps using a similar optimization procedure as in the main paper. 

\begin{figure*}[t]
    \centering
    \includegraphics[keepaspectratio,width=1.0\linewidth]{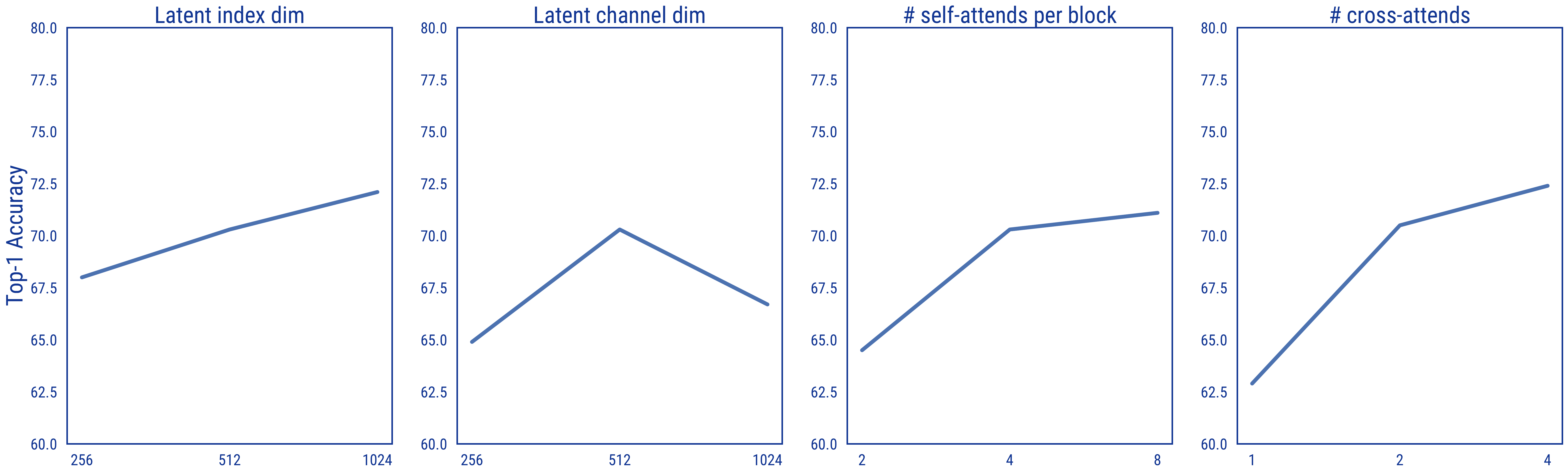}
    \vspace{-8pt}
    \caption{The effect of model hyperparameters, using a scaled-down Perceiver architecture on ImageNet. All plots show top-1 accuracy (higher is better). Increasing the size of the latent index dimension, the number of self-attends per block, and the number of cross-attends generally produced better results. Increasing the size of the latent channel dimension helps up to a point, but often leads to overfitting.}
    \label{fig:ablation}
\end{figure*}

The results from a hyperparameter sweep centered on this base architecture are shown in Fig.~\ref{fig:ablation}. All results show top-1 accuracy on ImageNet. Consistent with our other experiments, these results suggest that increasing the size of the model tends to produce better results. The exception in this experiment was the number of latent dimensions, as the largest model showed signs of overfitting.

Similarly, we evaluated the effect of the latent array's initialization scale and the parameters of the Fourier frequency position encoding on ImageNet performance. The results of this experiment are shown in Fig.~\ref{fig:init_ff}. These experiments use the full-sized ImageNet architecture, but were trained with a smaller batch size (256) and fewer TPUs (16) (for reasons of compute availability). These experiments suggest that standard and relatively small values for the initialization scale are best (values $\ge 1$ may lead to instability), and generally suggest that a higher number of Fourier frequency bands and a higher maximum resolution (up to Nyquist) improve performance. We found that a scale of 1.0 worked best for initializing the position encoding: this value is used for the model reported in Tab.~\ref{tab:imagenet_permuted}.

For all FLOPs numbers reported here, we report unfused multiply-adds

All FLOPS reported here give theoretical FLOPS with multiplies and accumulates counted as separate operations. This is the strategy used in \cite{kaplan2020scaling} and elsewhere in the literature. Note that some other papers in the literature report FLOPS using fused multiply-accumulates: using this strategy will approximately cut our reported figures in half.

\begin{table}[t]
\centering
\begin{tabular}{|l|l|l|l|}
\hline
\# cross-attends        & Acc. & FLOPs   & Params \\ \hline
4                       & 39.4 & 173.1B  & 12.7M   \\ 
8                       & 45.3 & 346.1B  & 23.8M   \\
12                      & OOM  & 519.2B  & 34.9M   \\ \hline
\end{tabular}
\vspace{-8pt}
\caption{Performance of models built from a stack of cross-attention layers with no latent transformers. We do not share weights between cross-attention modules in this experiment. Models with 12 cross-attends run out of memory on the largest device configuration we use (64 TPUs). Results are top-1 validation accuracy (in \%) on ImageNet (higher is better).}
\label{tab:no_transformers}
\end{table}

\begin{table}[t]
\centering
\begin{tabular}{|l|l|l|l|}
\hline
\# cross-attends        & Acc.          & FLOPs   & Params \\ \hline
1 (at start)            & 76.7          & 404.3B  & 41.1M   \\ 
1 (interleaved)         & 76.7          & 404.3B  & 42.1M   \\ \hline
2 (at start)            & 76.7          & 447.6B  & 44.9M   \\ 
2 (interleaved)         & 76.5          & 447.6B  & 44.9M   \\ \hline
4 (at start)            & 75.9          & 534.1B  & 44.9M   \\ 
4 (interleaved)         & 76.5          & 534.1B  & 44.9M   \\ \hline
8 (at start)            & 73.7          & 707.2B  & 44.9M   \\ 
8 (interleaved)         & \textbf{78.0} & 707.2B  & 44.9M   \\ \hline
\end{tabular}
\vspace{-8pt}
\caption{Performance as a function of \# of cross-attends and their arrangement. In ``interleaved,'' cross-attention layers are spaced throughout the network (for re-entrant processing), while in ``at start'' all cross-attends are placed at the start of the network followed by all latent self-attend layers. All cross-attention layers except the initial one are shared, and self-attends are shared as usual (using 8 blocks of 6 self-attention modules). Results are top-1 validation accuracy (in \%) on ImageNet (higher is better).}
\label{tab:cross_attend_config}
\vspace{-12pt}
\end{table}

\section{Architectural details}
\label{sec:supp_arch_details}

The Perceiver consists of two modules: a cross-attention module and a Transformer. In the cross-attention module, inputs are first processed with layer norm \cite{ba2016layer} before being passed through linear layers to produce each of the query, key, and value inputs to the QKV cross-attention operation. The queries, keys, and values have the same number of channels as the minimum of the input channels, which is typically the key/value input (i.e. 261 for ImageNet) The output of attention is passed through an additional linear layer to project it to the same number of channels in the query inputs (so it can be added residually).

\begin{figure*}[t]
    \centering
    \includegraphics[keepaspectratio,width=0.9\linewidth]{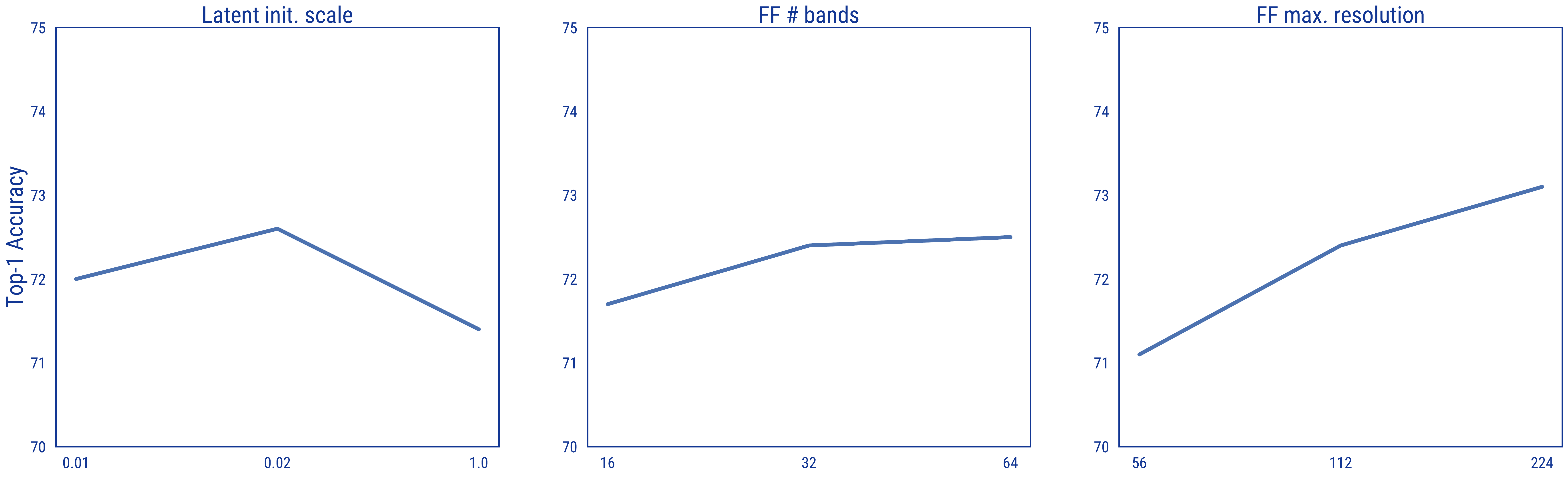}
    \vspace{-12pt}
    \caption{The effect of latent initialization scale and Fourier feature (FF) position encoding parameters on performance. All plots show top-1 accuracy (higher is better). The model with initialization scale of 0.1 diverged during training. Generally, increasing the number of bands and max resolution (up to Nyquist) increased performance. We observed the same effects whether using linearly or logarithmically spaced position encoding bands.}
    \label{fig:init_ff}
\end{figure*}

The query inputs for the first cross-attention layer (e.g. the left-most latent array in Fig.~\ref{fig:architecture}) are learned, per-element weights with the same shape as the latent array (e.g. for ImageNet, they are a $512 \times 1024$ array). These function like learned position encodings in the Transformer literature or like a learned initial state in the recurrent neural network (RNN) literature. The latent array is randomly initialized using a truncated normal distribution with mean 0, standard deviation 0.02, and truncation bounds [-2, 2]. Network performance is fairly robust to the scale of this initialization (see Fig.~\ref{fig:init_ff}, left facet).

In the self-attention block, inputs are processed with layer norm and passed through query, key, and value layers before being used to compute QKV self-attention. The output is passed through another linear layer.

Each cross-attention and self-attention block is followed by a dense (multi-layer Perceptron) block. In the dense block, inputs are processed with layer norm, passed through a linear layer, activated with a GELU nonlinearity \cite{hendrycks2016gelu}, and passed through a final linear layer. We used dropout throughout the network in earlier experiments, but we found this led to degraded performance, so no dropout is used.

All linear layers (including query, key, and value layers and dense block layers) preserve the dimensionality of their inputs and are tiled over input index dimensions (i.e. applied as a 1 $\times$ 1 convolution).

To produce output logits, we average the output of the final latent self-attention module over the index dimension. This produces a single global summary vector, which we then project to the number of target classes using a single linear layer. This is the process used by e.g. ResNets to convert a convolutional feature map to logits \cite{he2016deep}.

As with other Transformer architectures, the Perceiver's Transformer has a fully residual design, and its input is always added to its output for further processing. This applies to cross-attention modules as well: the latent component of the input is added to its output. We give details on the hyperparameters used on different datasets in the main paper.

\section{Position encodings and Fourier features}

\noindent \textbf{Crop-relative coordinates.} As described in the main paper, we found that generating position coordinates using cropped data rather than on the raw data was important to prevent excessive overfitting. We illustrate the cropping procedure on ImageNet in Fig.~\ref{fig:cropping}.

\begin{table}[]
\begin{tabular}{|l|l|l|l|l|}
                                                                      \hline
                  & Valid         & Train       & Params & FLOPs   \\ \hline
No weight sharing & 72.9          & 87.7        & 326.2M & 707.2B  \\ \hline
W/ weight sharing & \textbf{78.0} & 79.5        & 44.9M  & 707.2B  \\ \hline
\end{tabular}
\caption{Weight sharing mitigates overfitting and leads to better validation performance on ImageNet. We show results (top-1 accuracy) for the best-performing ImageNet architecture (reported in Tables 1-2 of the main paper) on train and validation sets. This architecture uses 8 cross-attends and 6 blocks per latent Transformer. The model labeled ``W/ weight sharing'' shares weights between cross-attention modules 2-8 and between the corresponding blocks of all latent Transformers. The first cross-attention module uses its own, unshared weights.}
\label{tab:weight_sharing}
\vspace{-12pt}
\end{table}

\noindent \textbf{Fourier feature parameterizations.} We choose the Fourier feature parameterization described in section 3.2 of the paper to allow us to intuitively set the maximum band when the sample rate of the input signal is regular and known. By setting the number of bands independently, we allow it be easily controlled in line with a computational budget: we generally found that more bands helped for a given architecture (assuming it fits in memory). For signals with irregular or very fine sampling, such as ModelNet40 point clouds, the maximum band can also be treated as a hyperparameter. This is in contrast to the parameterization used in NeRF \cite{mildenhall2020nerf}, which produces very high frequencies if a moderate number of bands are used (e.g. the $64^{\text{th}}$ band would have a frequency of $2^{64}=1.8e19$). Rather than tying the maximum frequency to the number of bands, our parameterization samples the spectrum more densely as more bands are added. Our parameterization is identical to the parameterization described in the original Transformer paper, except we express each band in terms of its frequency rather than its wavelength (we find this more natural in the context of signals like images) and we assume that input positions are in $[-1, 1]$ rather than $[0, s)$ for a sequence of length $s$.


\section{Audiovisual attention maps}
\label{sec:audiovisual_viz}

We visualize video and audio attention maps (respectively) for the first and second cross-attention module of a multimodal Perceiver model trained on AudioSet using 2x4x4 video patches and 61,440 raw audio samples 

We visualize video attention maps similarly to static image attention maps (Fig.~\ref{fig:attention_map}), but with the addition of a time dimension: each column shows the attention to the full image at a time step of the video. Because this AudioSet Perceiver takes space-time patches of shape time 2 $\times$ height 4 $\times$ width 4, the same attention is applied to pairs of subsequent frames. For visualization purposes, we show every other frame of the input video and attention maps: each attention map is applied to two video frames.

All attention maps of this network appear to be sensitive to both static and dynamic features of the input video. Some attention maps exhibit spatiotemporal structure reminiscent of the filters seen in spatiotemporal image processing \cite{adelson1985spatiotemporal, simoncelli1998model}. Because the Perceiver uses learned attention rather than a fixed bank of spatiotemporal filters, it can adapt its attention to the input content.

We visualize audio attention maps by displaying the mel-spectrogram and attention maps as images. Mel-spectrograms are plotted with time plotted on the x- and frequency on the y-axis. Although they are harder to interpret visually than the image attention maps, they appear to share a common structure of Fourier-frequency positional banding and content-related modulation.

\section{Notes on changes from the original version}

Our Audioset mAP results in the first arXiv version were flawed (and unfortunately higher) so we repeated and expanded those experiments and now provide correct numbers. The issue was that when computing AP using the sklearn package, we passed the matrix of class scores transposed to what the function expects -- hence the number of classes and number of examples were switched.

We have slightly improved the results reported on ImageNet since the first version by (i) removing dropout, (ii) removing a linear layer that was originally (unintentionally) included following the initial latent array, and (iii) averaging before rather than after projecting when computing the output logits.

\begin{figure*}
    \centering
    \includegraphics[keepaspectratio,width=1.0\linewidth]{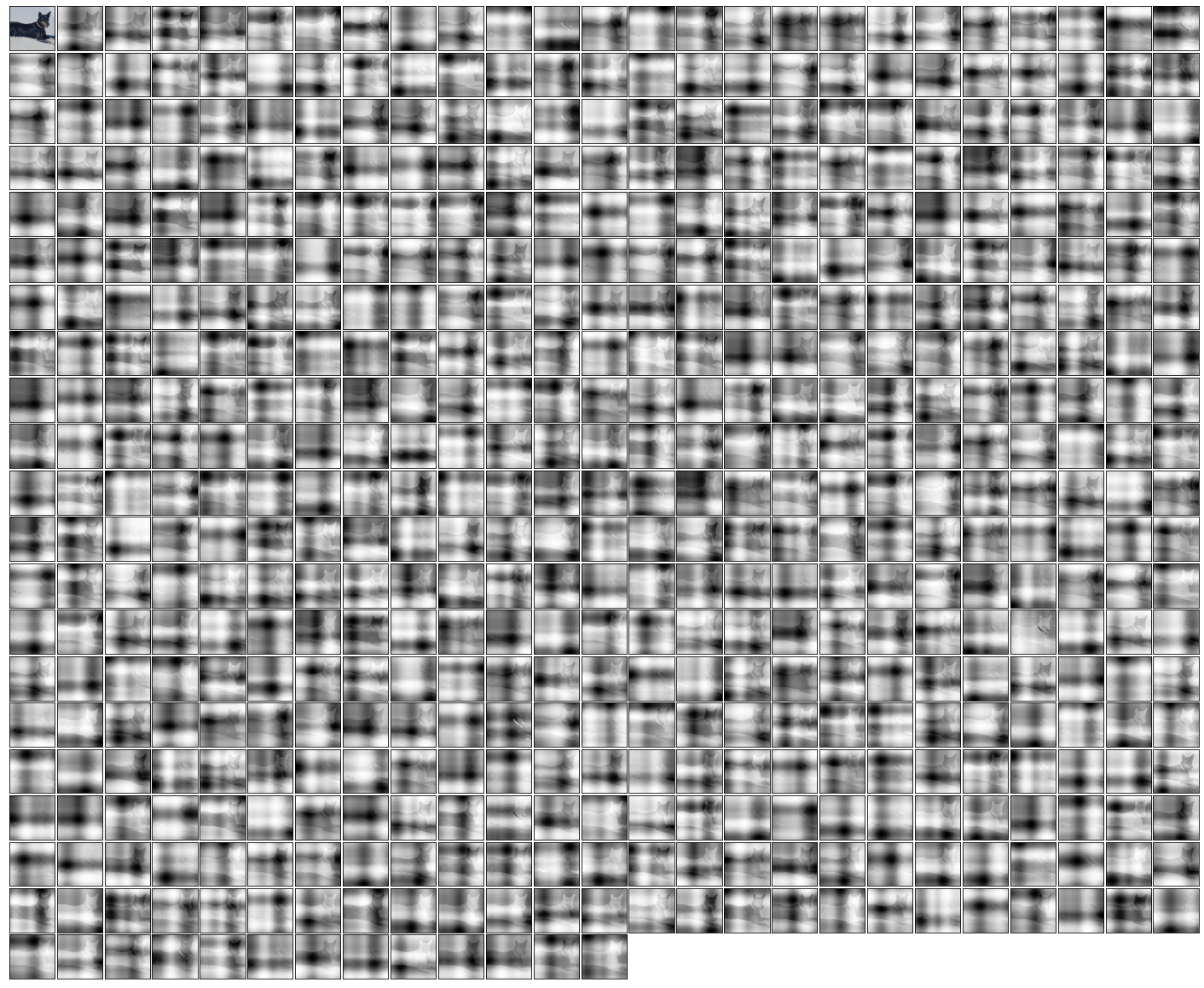}
    \vspace{-12pt}
    \caption{Example attention maps from the \textbf{first cross-attend} of an ImageNet network trained with \textbf{2D Fourier feature} position encodings.}
\end{figure*}

\begin{figure*}
    \centering
    \includegraphics[keepaspectratio,width=1.0\linewidth]{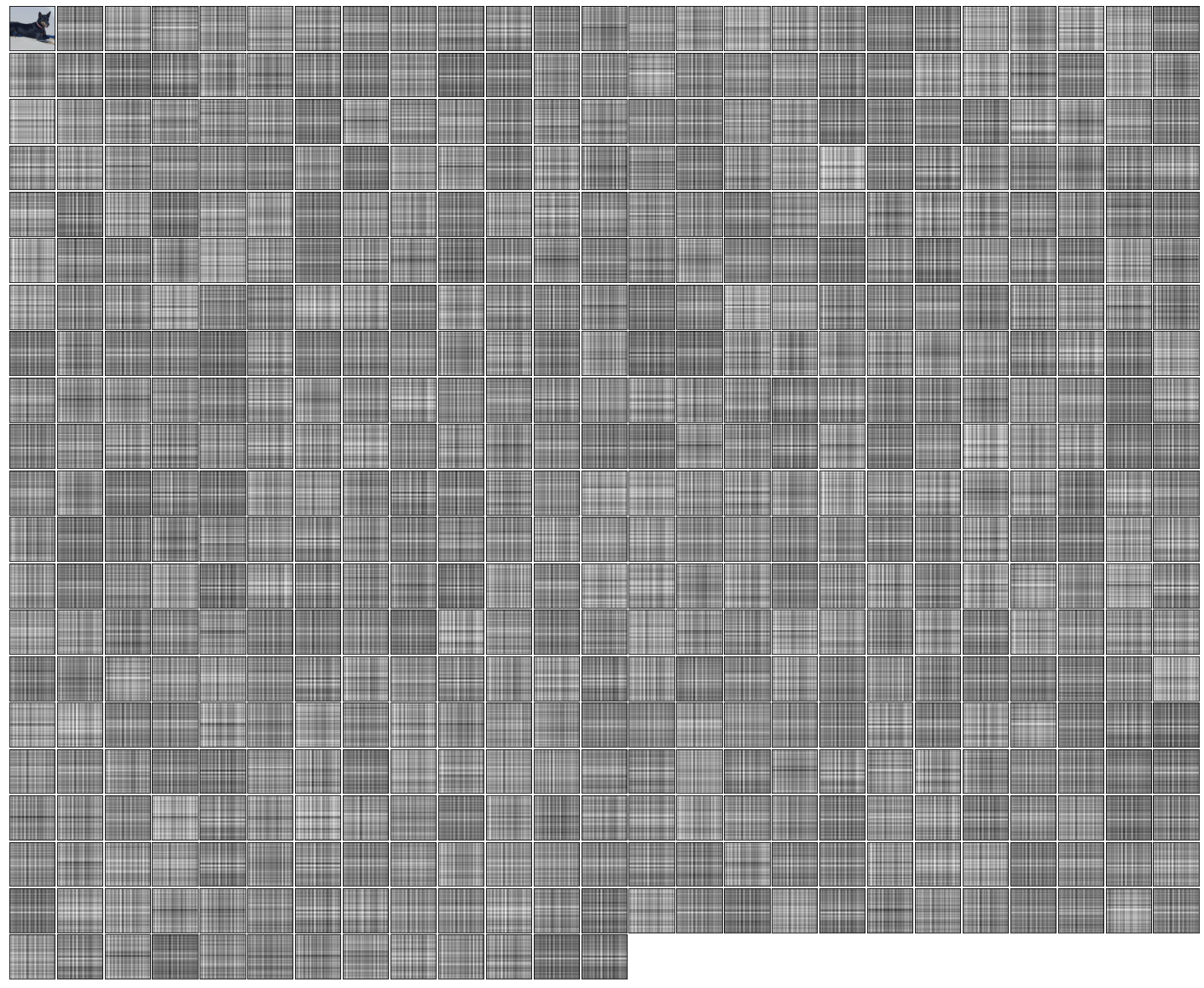}
    \vspace{-12pt}
    \caption{Example attention maps from the \textbf{eighth (final) cross-attend} of an ImageNet network trained with \textbf{2D Fourier feature} position encodings.}
\end{figure*}

\begin{figure*}
    \centering
    \includegraphics[keepaspectratio,width=1.0\linewidth]{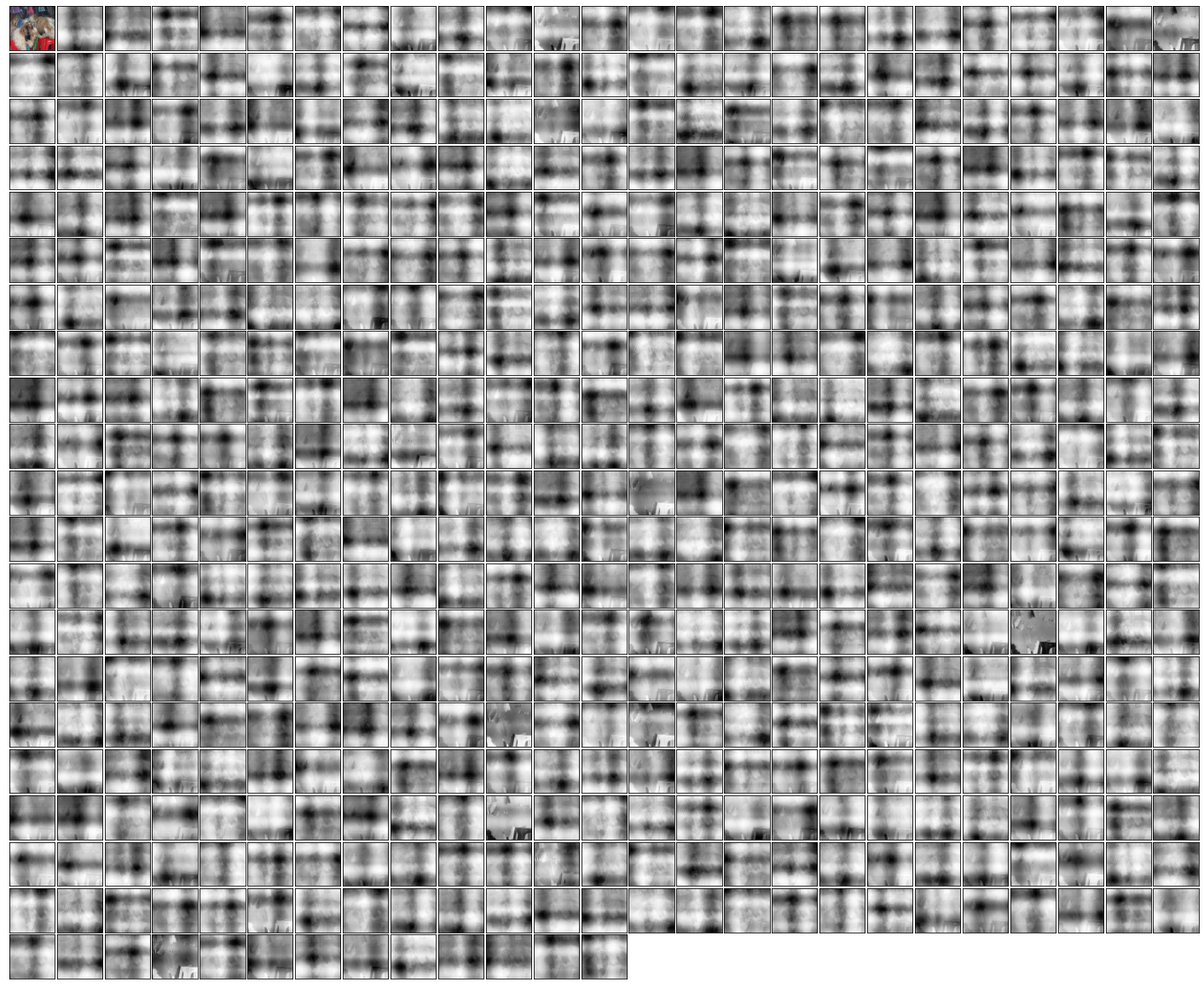}
    \vspace{-12pt}
    \caption{Example attention maps from the \textbf{first cross-attend} of an ImageNet network trained with \textbf{2D Fourier feature} position encodings.}
\end{figure*}

\begin{figure*}
    \centering
    \includegraphics[keepaspectratio,width=1.0\linewidth]{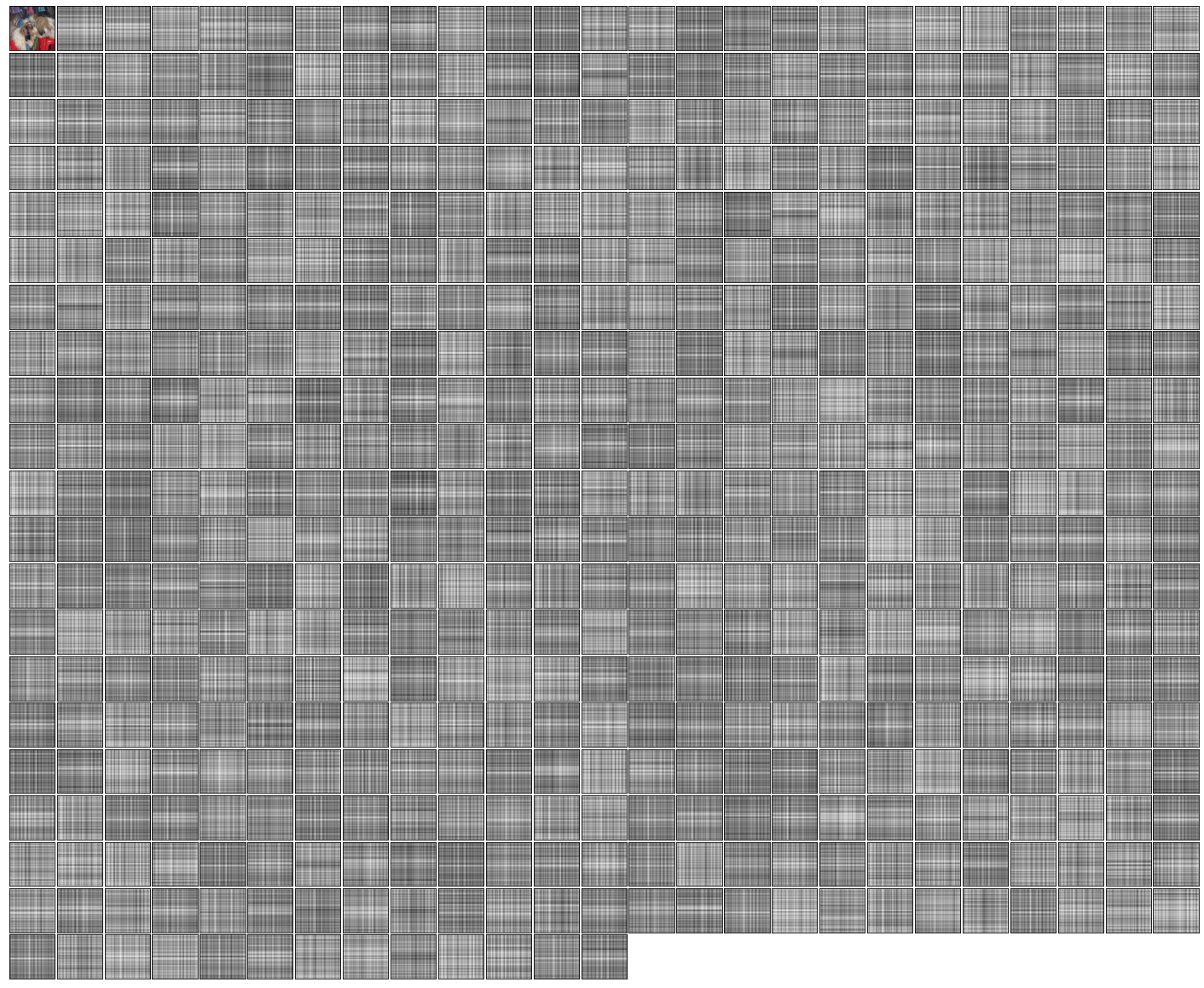}
    \vspace{-12pt}
    \caption{Example attention maps from the \textbf{eighth (final) cross-attend} of an ImageNet network trained with \textbf{2D Fourier feature} position encodings.}
\end{figure*}

\begin{figure*}
    \centering
    \includegraphics[keepaspectratio,width=1.0\linewidth]{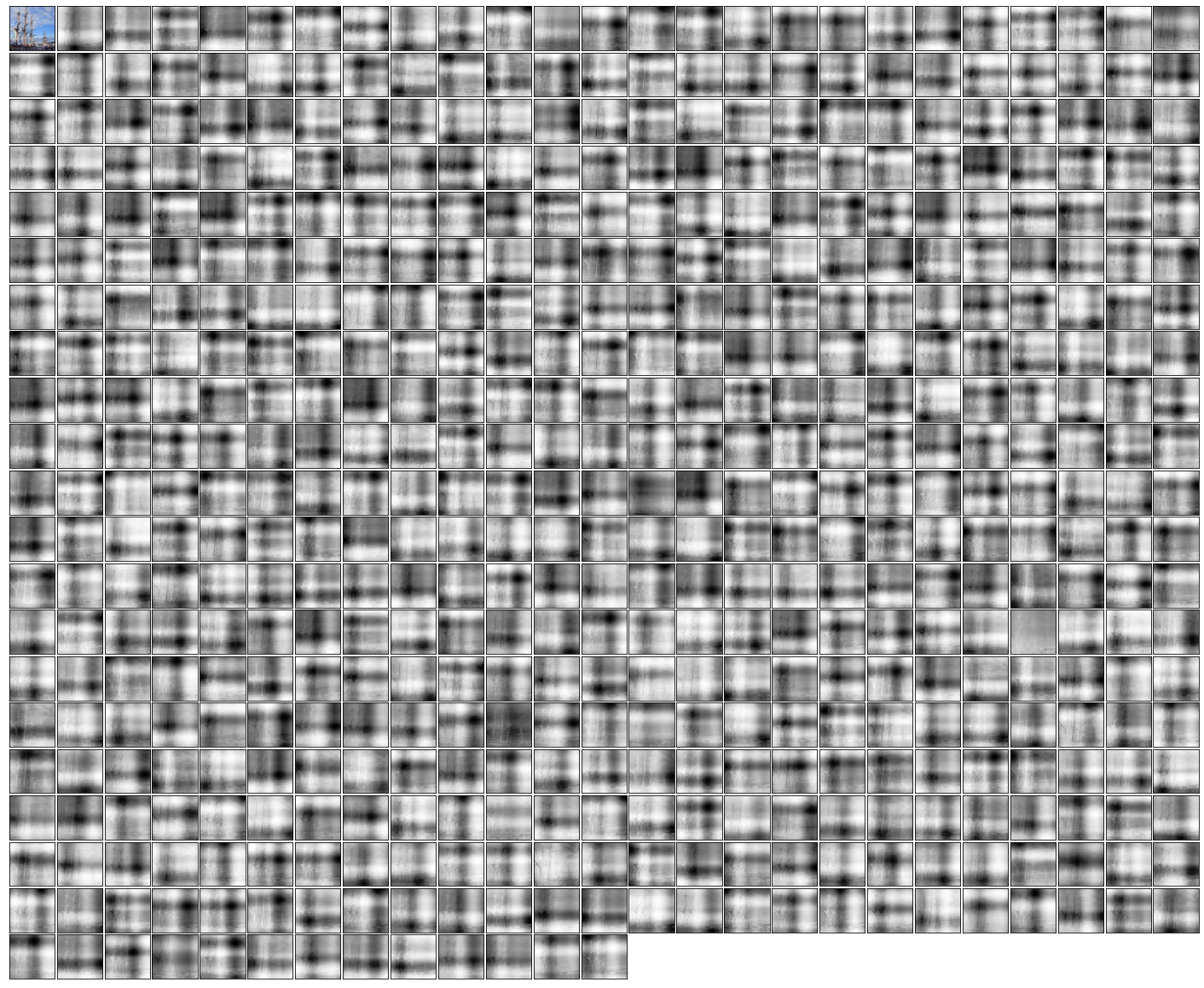}
    \vspace{-12pt}
    \caption{Example attention maps from the \textbf{first cross-attend} of an ImageNet network trained with \textbf{2D Fourier feature} position encodings.}
\end{figure*}

\begin{figure*}
    \centering
    \includegraphics[keepaspectratio,width=1.0\linewidth]{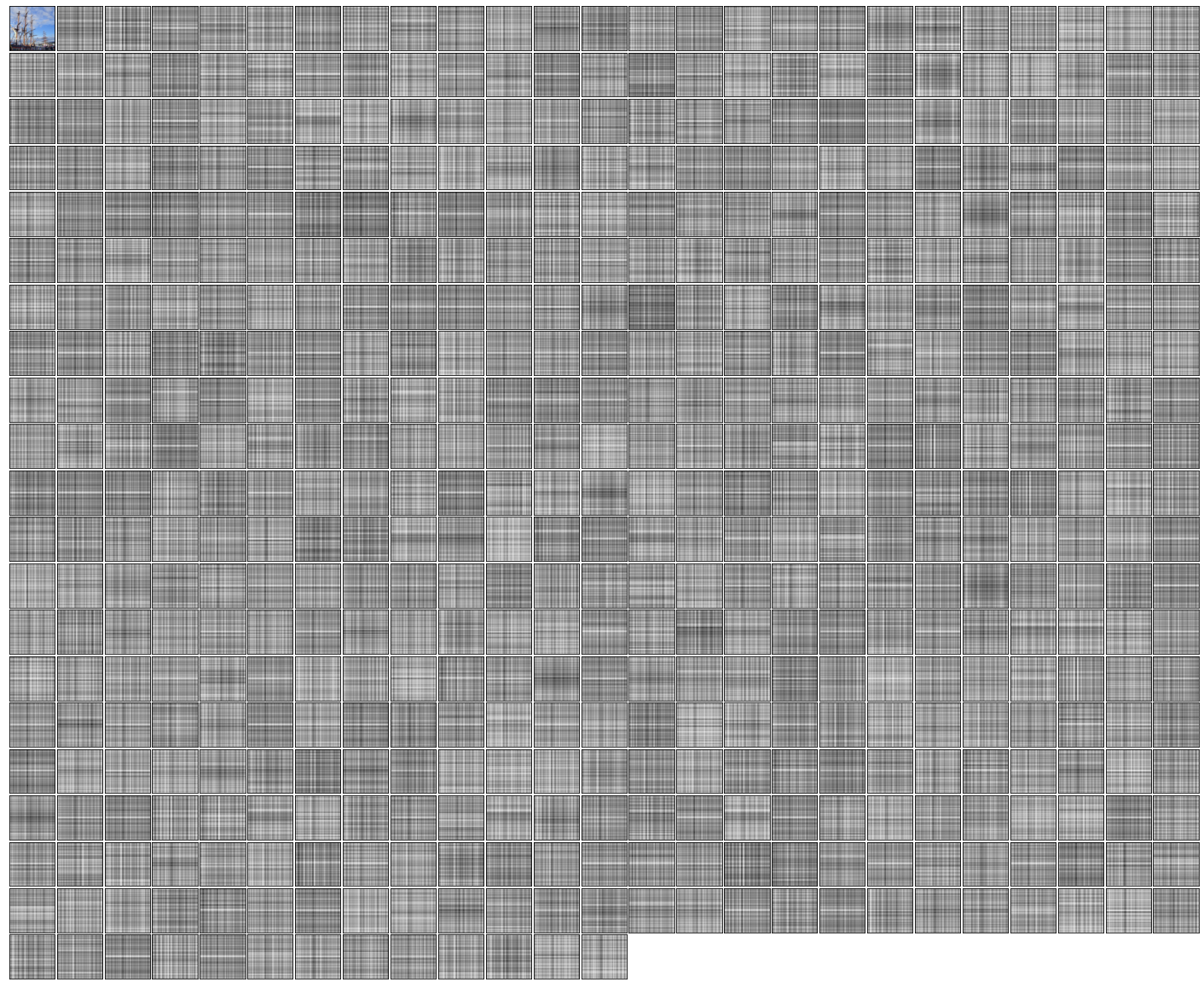}
    \vspace{-12pt}
    \caption{Example attention maps from the \textbf{eighth (final) cross-attend} of an ImageNet network trained with \textbf{2D Fourier feature} position encodings.}
\end{figure*}

\begin{figure*}
    \centering
    \includegraphics[keepaspectratio,width=1.0\linewidth]{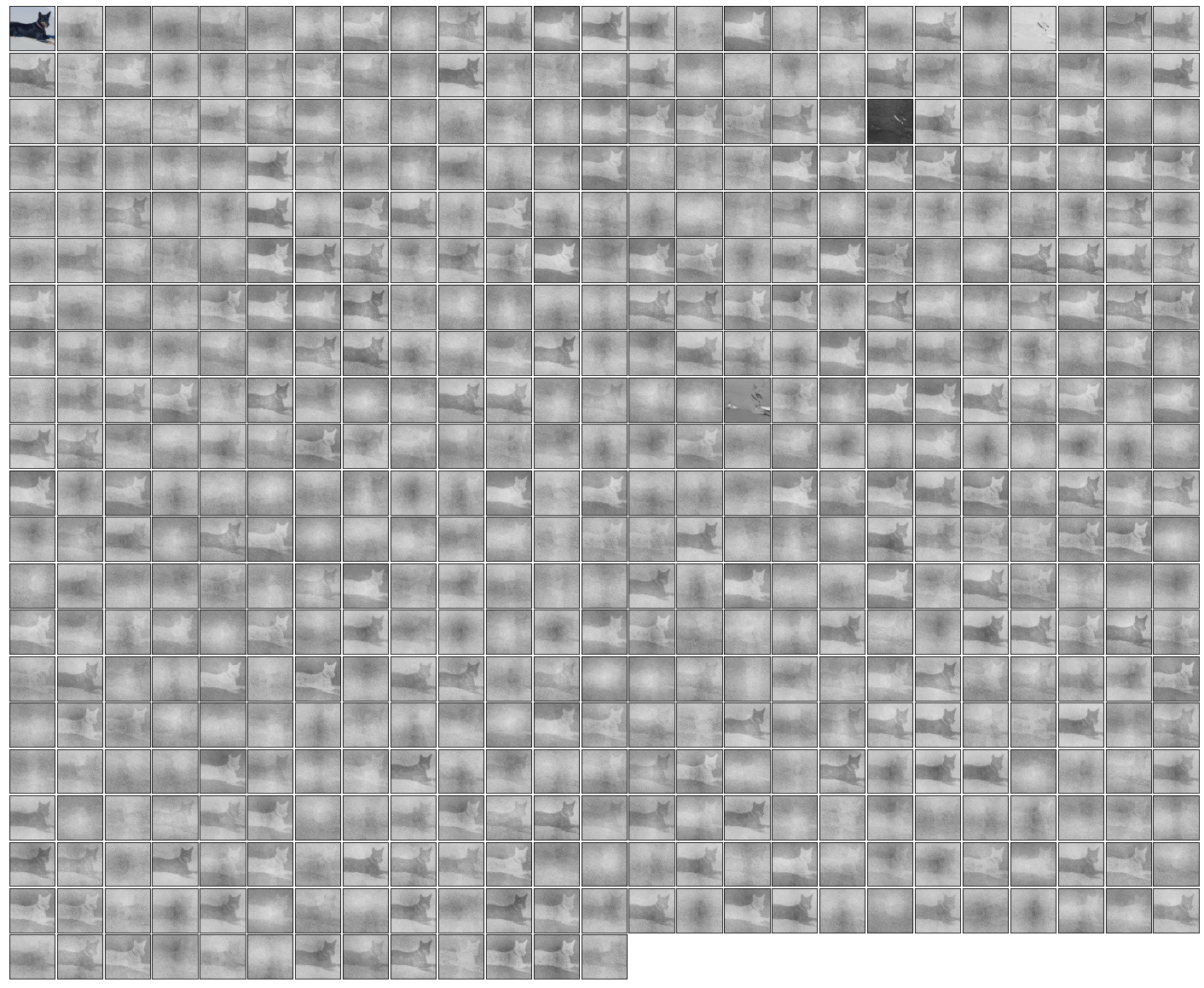}
    \vspace{-12pt}
    \caption{Example attention maps from the \textbf{first (only) cross-attention} of an ImageNet network trained with \textbf{learned position encodings}.}
\end{figure*}

\begin{figure*}
    \centering
    \includegraphics[keepaspectratio,width=1.0\linewidth]{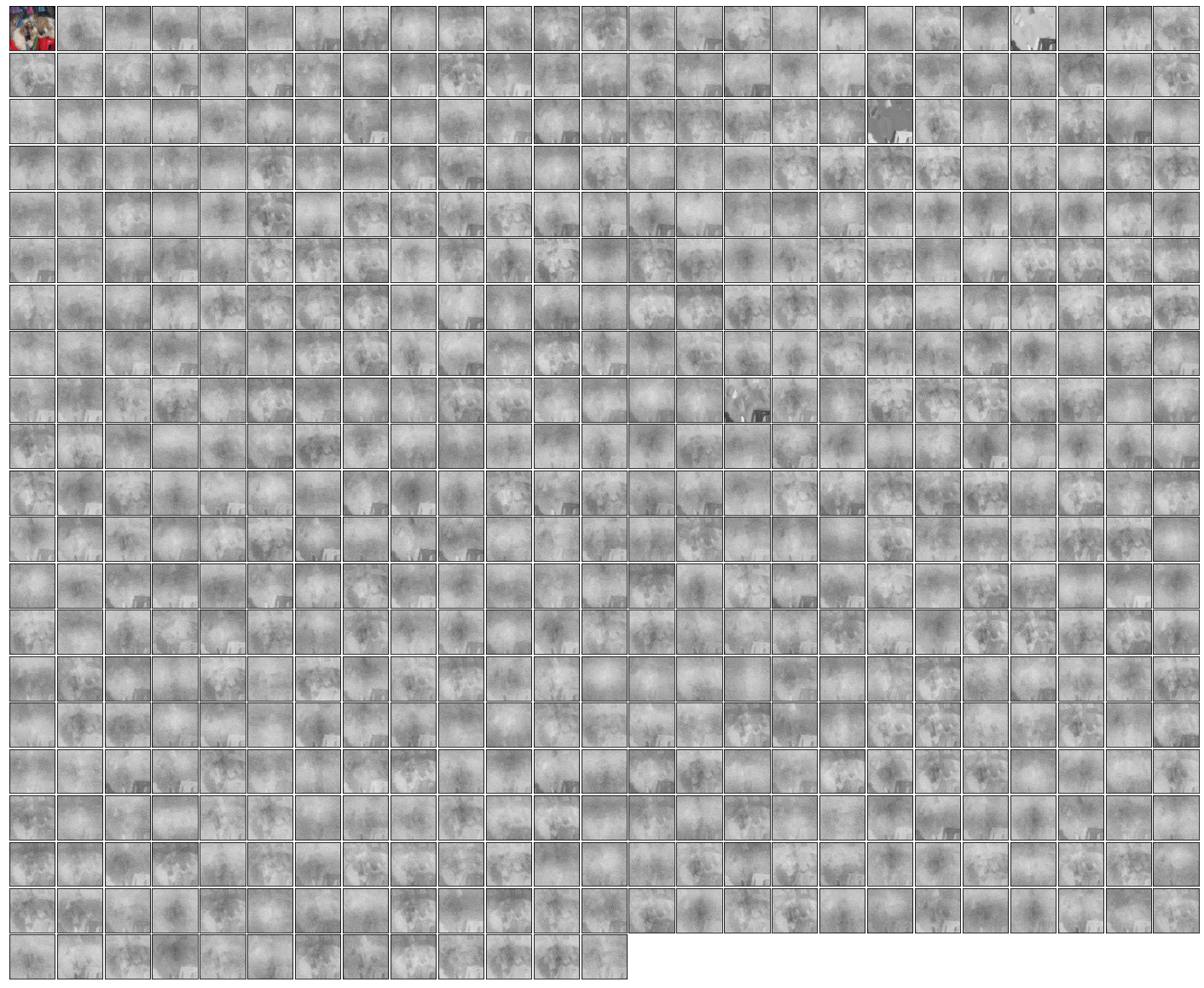}
    \vspace{-12pt}
    \caption{Example attention maps from the \textbf{first (only) cross-attention} of an ImageNet network trained with \textbf{learned position encodings}.}
\end{figure*}

\begin{figure*}
    \centering
    \includegraphics[keepaspectratio,width=1.0\linewidth]{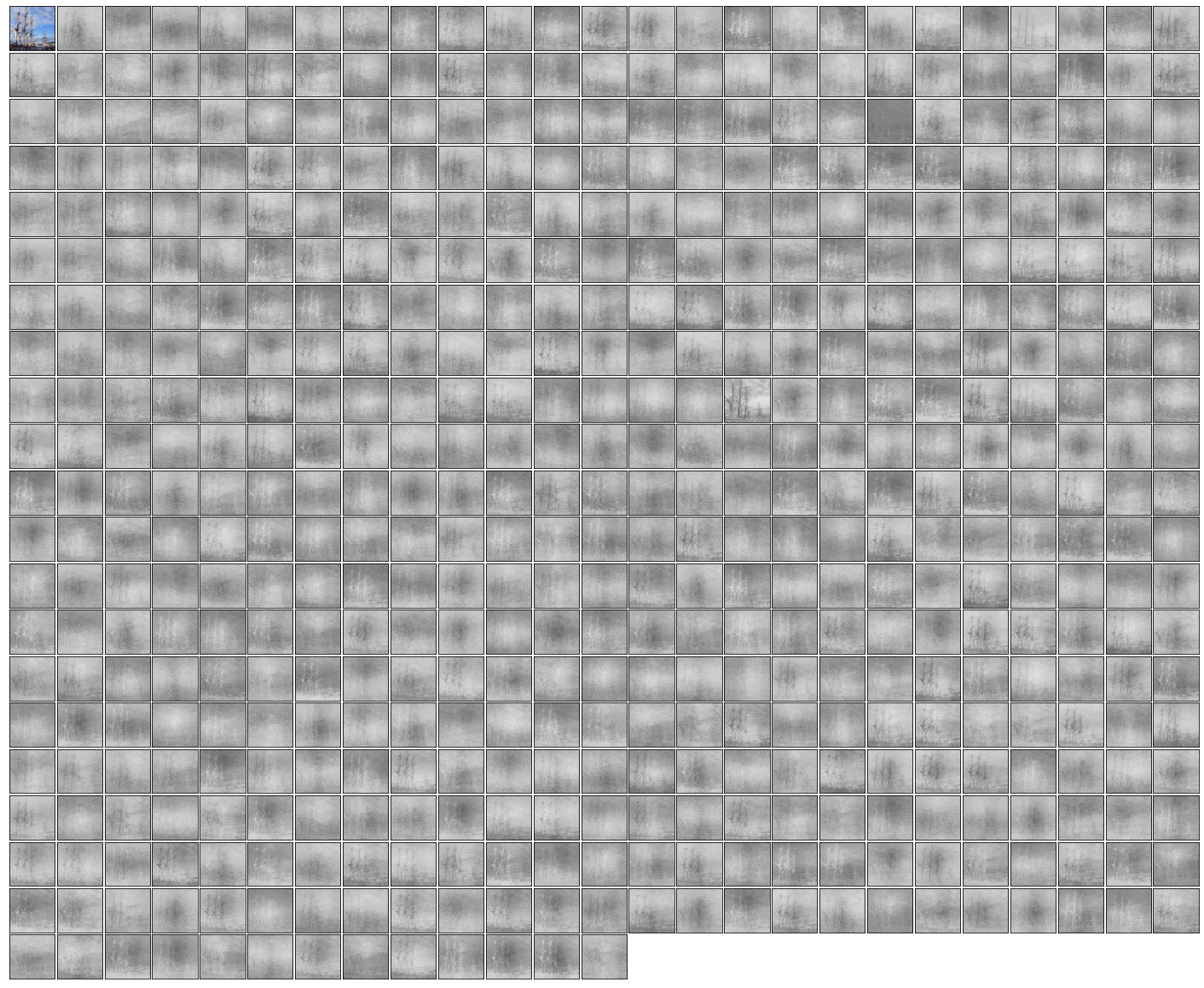}
    \vspace{-12pt}
    \caption{Example attention maps from the \textbf{first (only) cross-attention} of an ImageNet network trained with \textbf{learned position encodings}.}
\end{figure*}

\begin{figure*}
    \centering
    \includegraphics[keepaspectratio,width=1.0\linewidth]{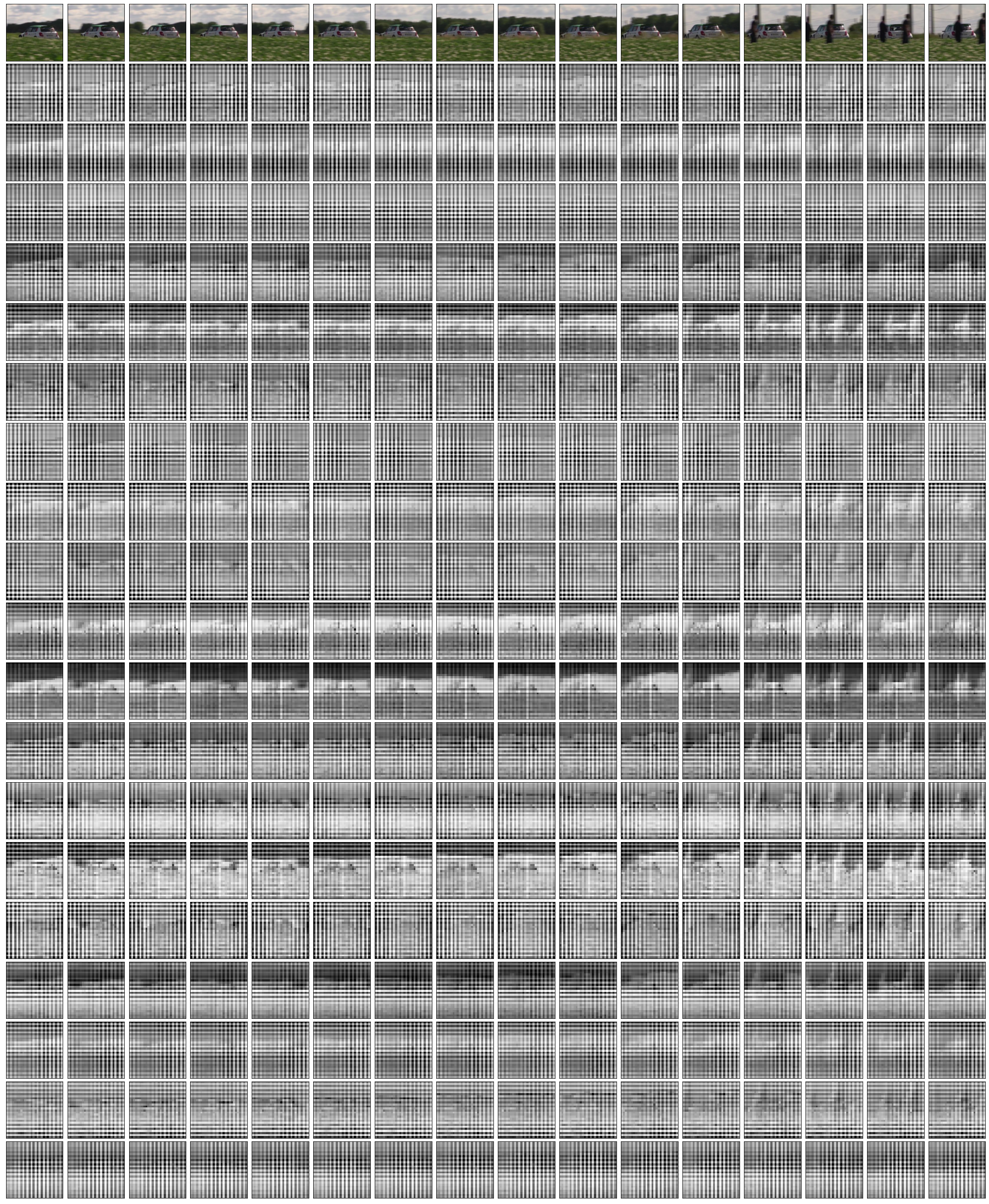}
    \vspace{-12pt}
    \caption{Example attention maps from the \textbf{first cross-attend} of an AudioSet network trained on \textbf{video only}.}
\end{figure*}

\begin{figure*}
    \centering
    \includegraphics[keepaspectratio,width=1.0\linewidth]{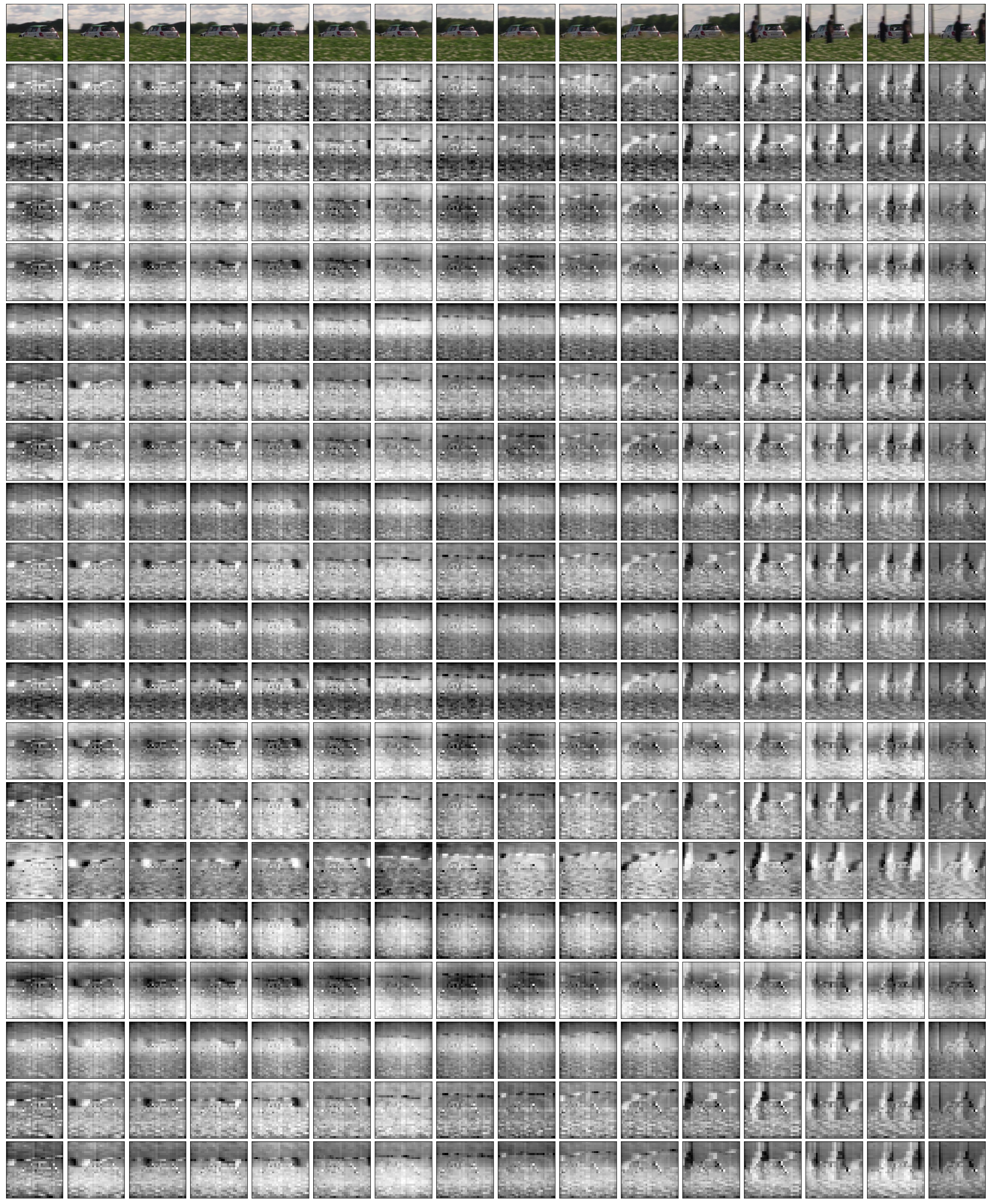}
    \vspace{-12pt}
    \caption{Example attention maps from the \textbf{second (final) cross-attend} of an AudioSet network trained on \textbf{video only}.}
\end{figure*}

\begin{figure*}
    \centering
    \includegraphics[keepaspectratio,width=1.0\linewidth]{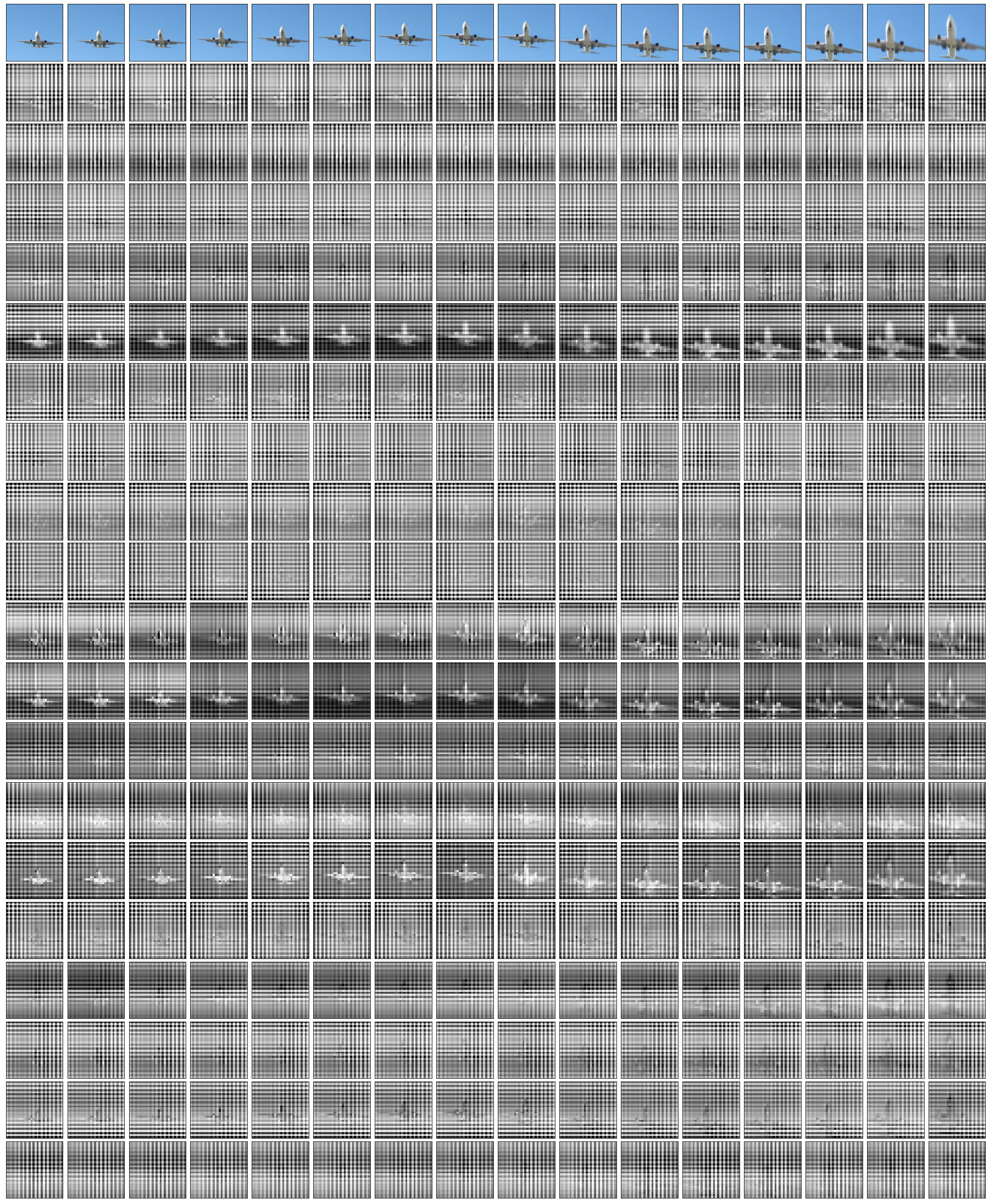}
    \vspace{-12pt}
    \caption{Example attention maps from the \textbf{first cross-attend} of an AudioSet network trained on \textbf{video only}.}
\end{figure*}

\begin{figure*}
    \centering
    \includegraphics[keepaspectratio,width=1.0\linewidth]{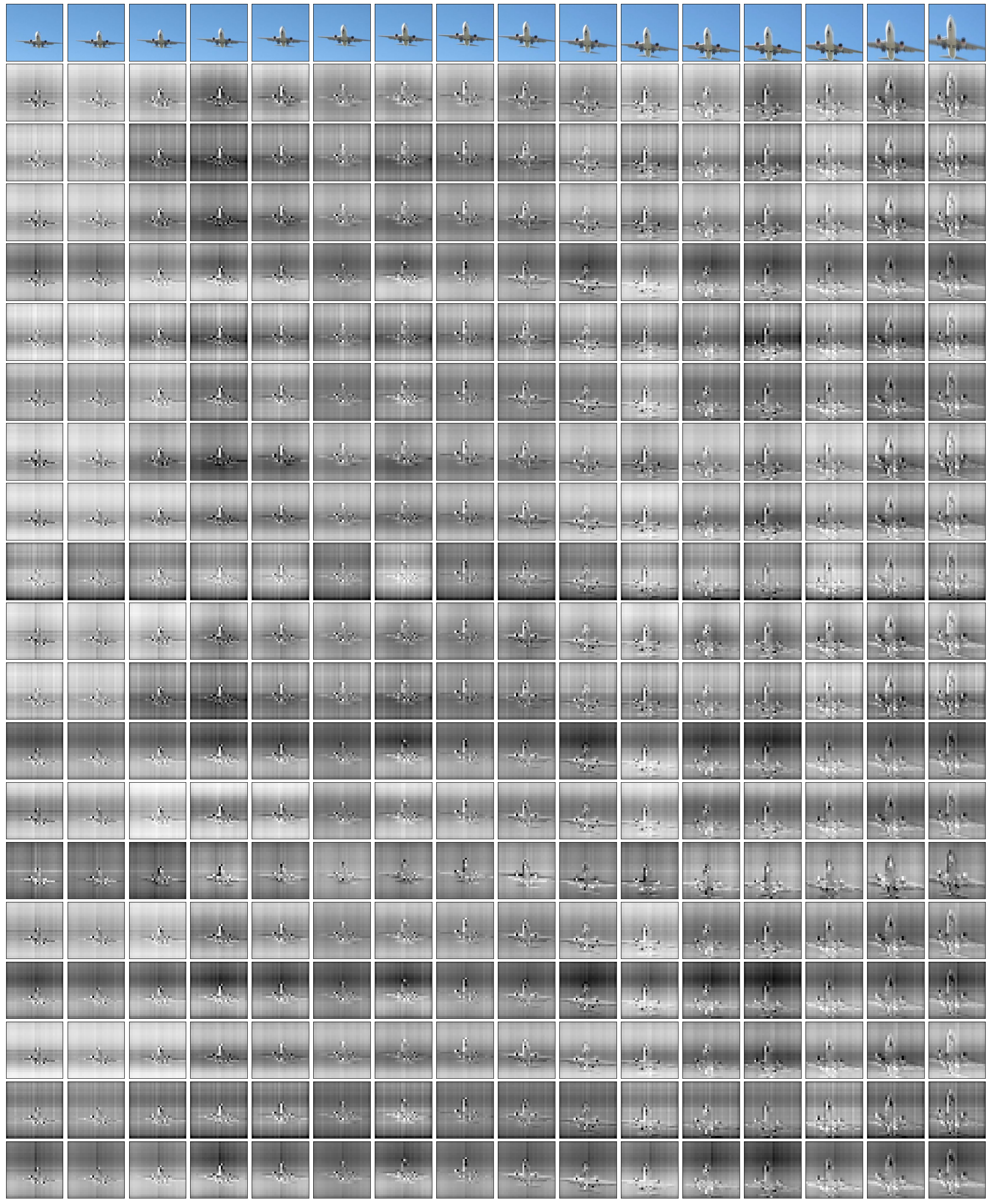}
    \vspace{-12pt}
    \caption{Example attention maps from the \textbf{second (final) cross-attend} of an AudioSet network trained on \textbf{video only}.}
\end{figure*}

\begin{figure*}
    \centering
    \includegraphics[keepaspectratio,width=1.0\linewidth]{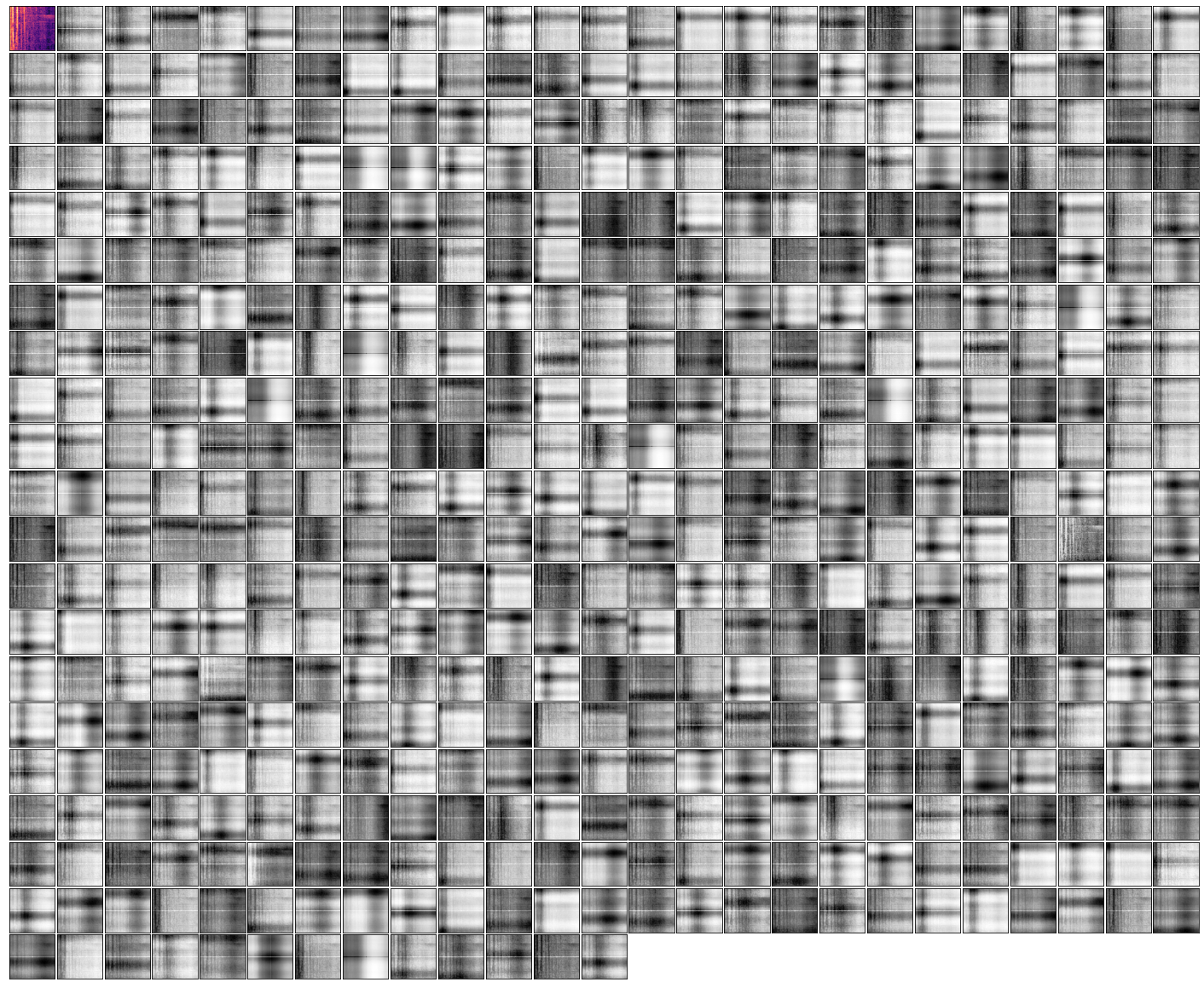}
    \vspace{-12pt}
    \caption{Example attention maps from the \textbf{first cross-attend} of an AudioSet network trained on \textbf{mel-spectrogram only} (car).}
\end{figure*}

\begin{figure*}
    \centering
    \includegraphics[keepaspectratio,width=1.0\linewidth]{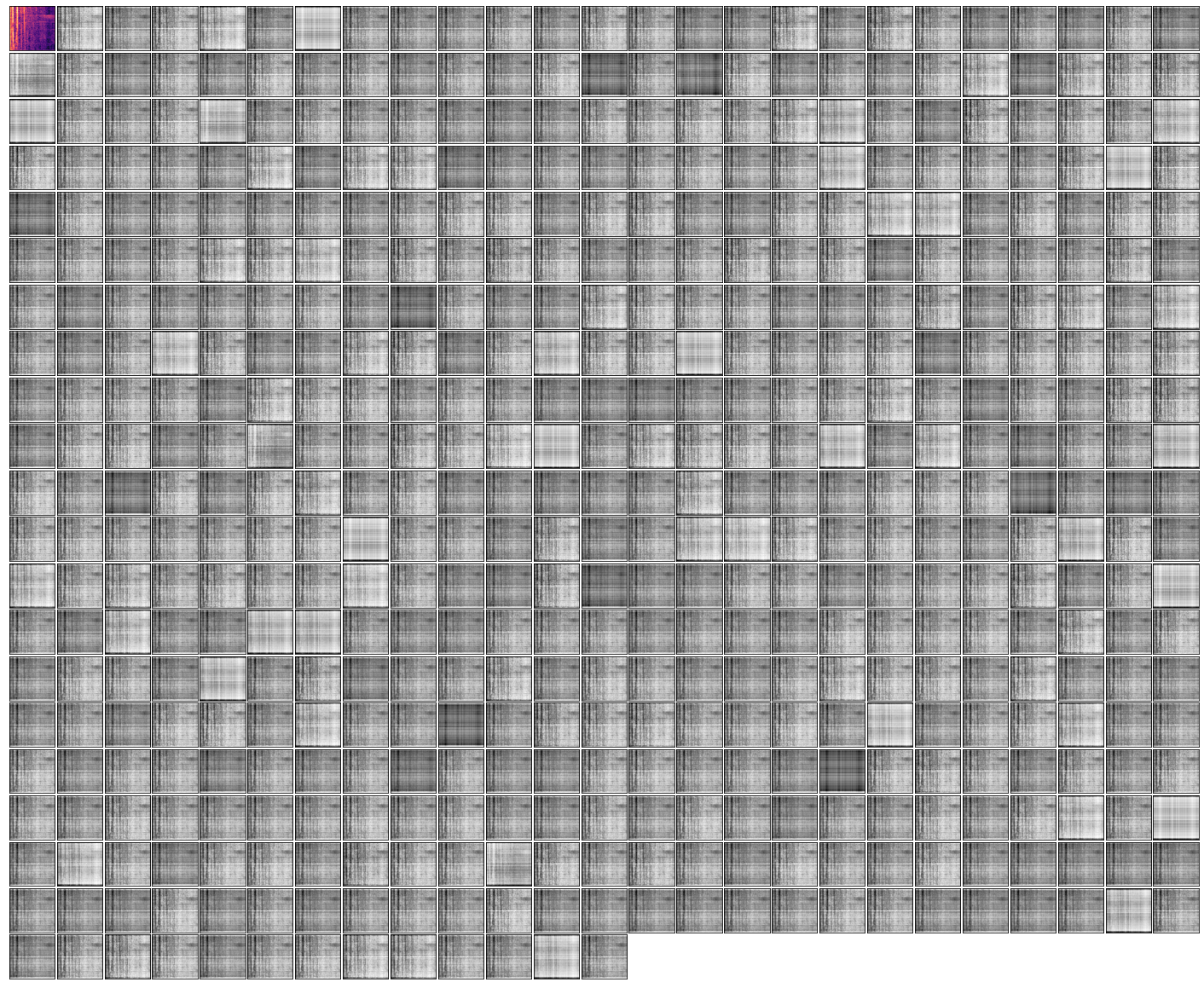}
    \vspace{-12pt}
    \caption{Example attention maps from the \textbf{second (final) cross-attend} of an AudioSet network trained on \textbf{mel-spectrogram only} (car).}
\end{figure*}

\begin{figure*}
    \centering
    \includegraphics[keepaspectratio,width=1.0\linewidth]{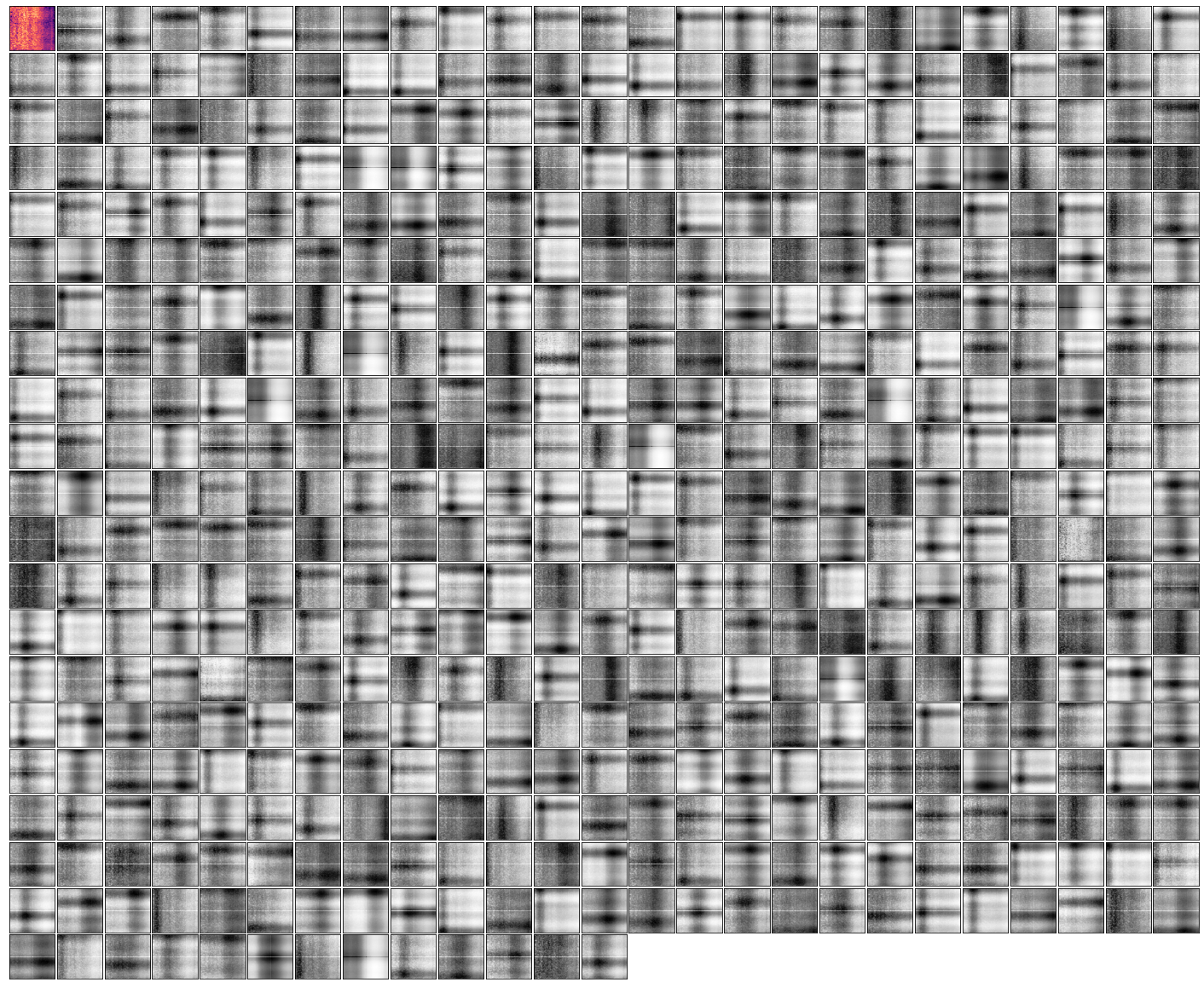}
    \vspace{-12pt}
    \caption{Example attention maps from the \textbf{first cross-attend} of an AudioSet network trained on \textbf{mel-spectrogram only} (plane).}
\end{figure*}

\begin{figure*}
    \centering
    \includegraphics[keepaspectratio,width=1.0\linewidth]{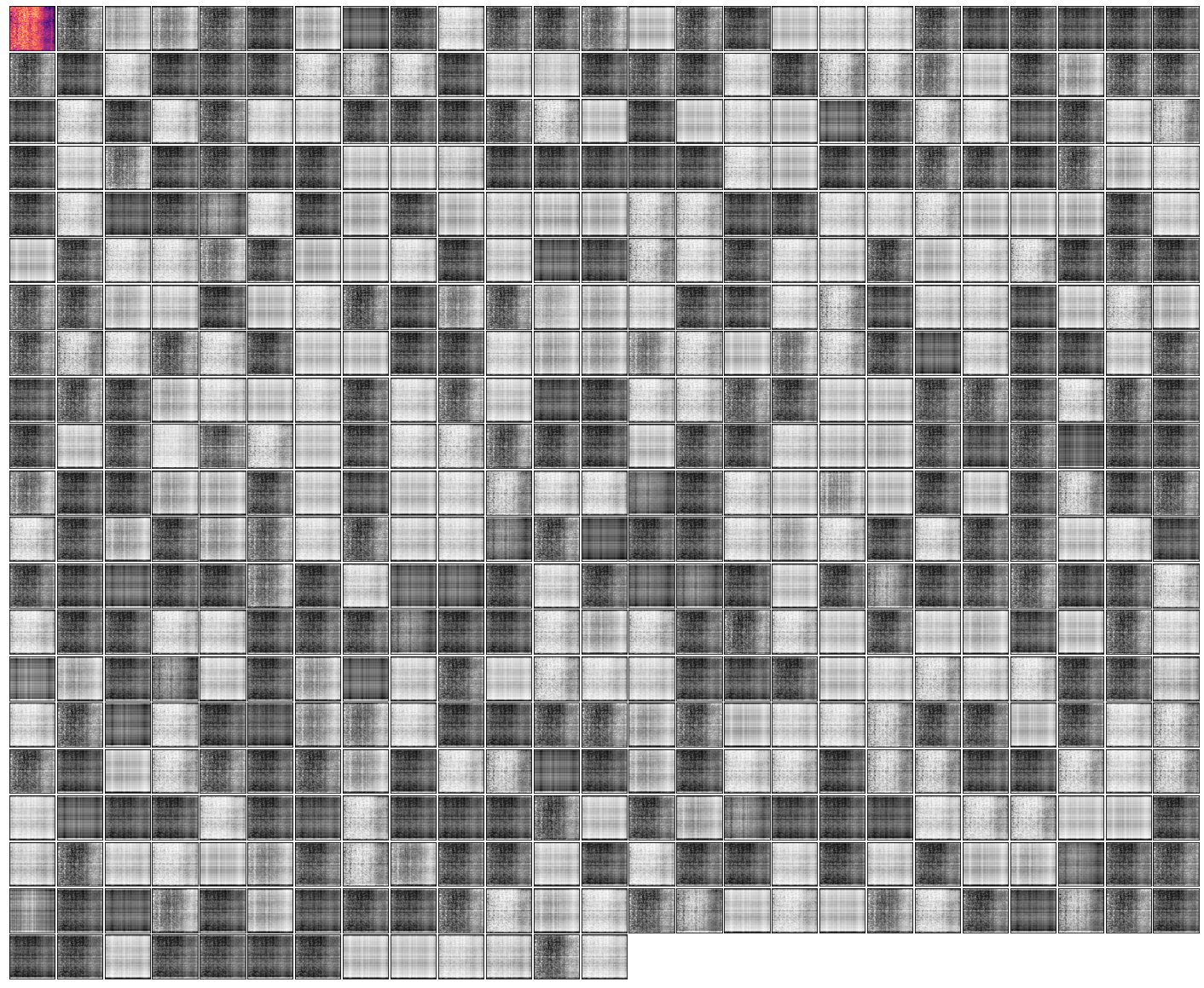}
    \vspace{-12pt}
    \caption{Example attention maps from the \textbf{second (final) cross-attend} of an AudioSet network trained on \textbf{mel-spectrogram only} (plane).}
\end{figure*}

\begin{figure*}
    \centering
    \includegraphics[keepaspectratio,width=1.0\linewidth]{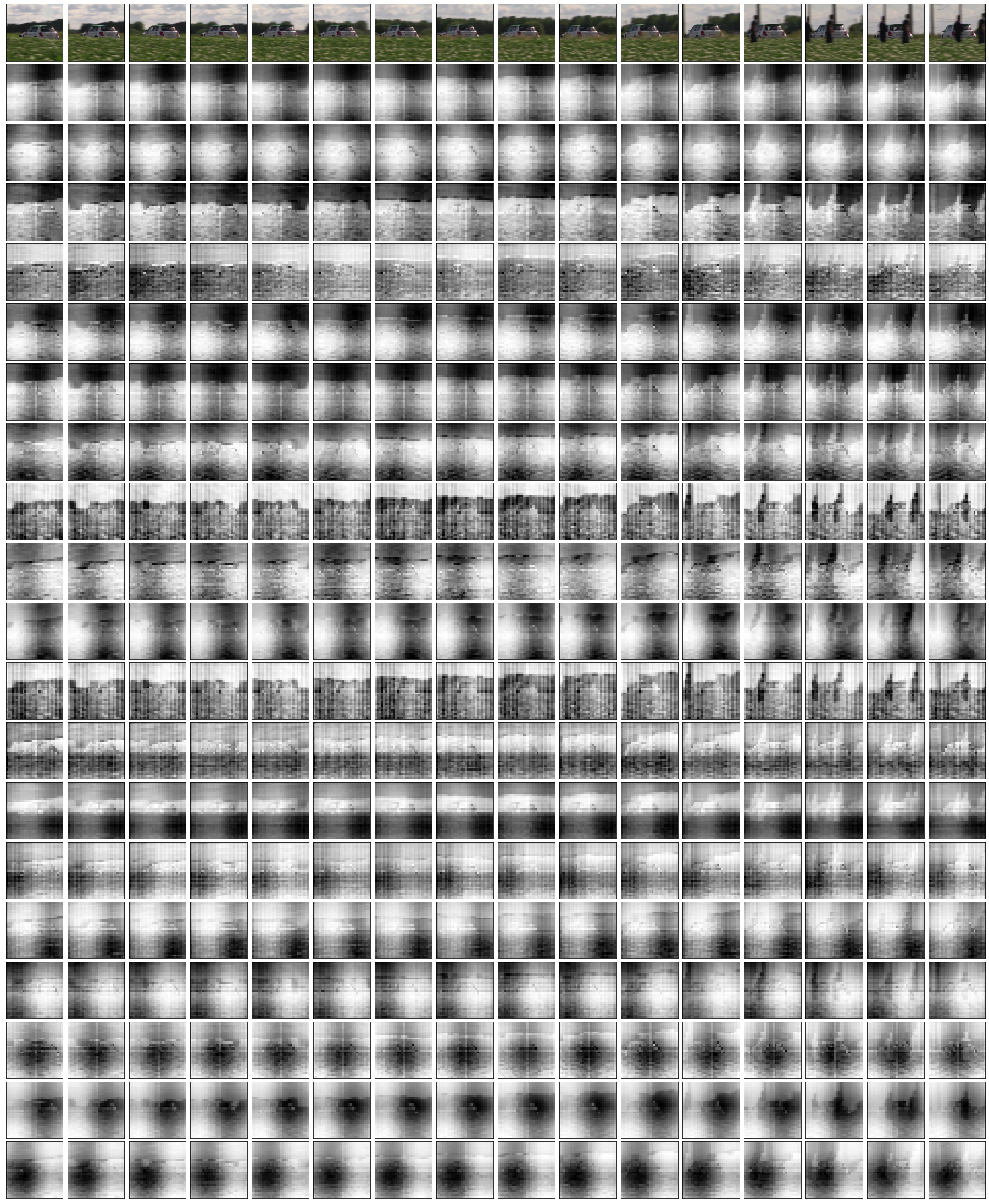}
    \vspace{-12pt}
    \caption{Example attention maps from the \textbf{first cross-attend} over the video input subset of an AudioSet network trained on \textbf{video and mel-spectrogram}.}
\end{figure*}

\begin{figure*}
    \centering
    \includegraphics[keepaspectratio,width=1.0\linewidth]{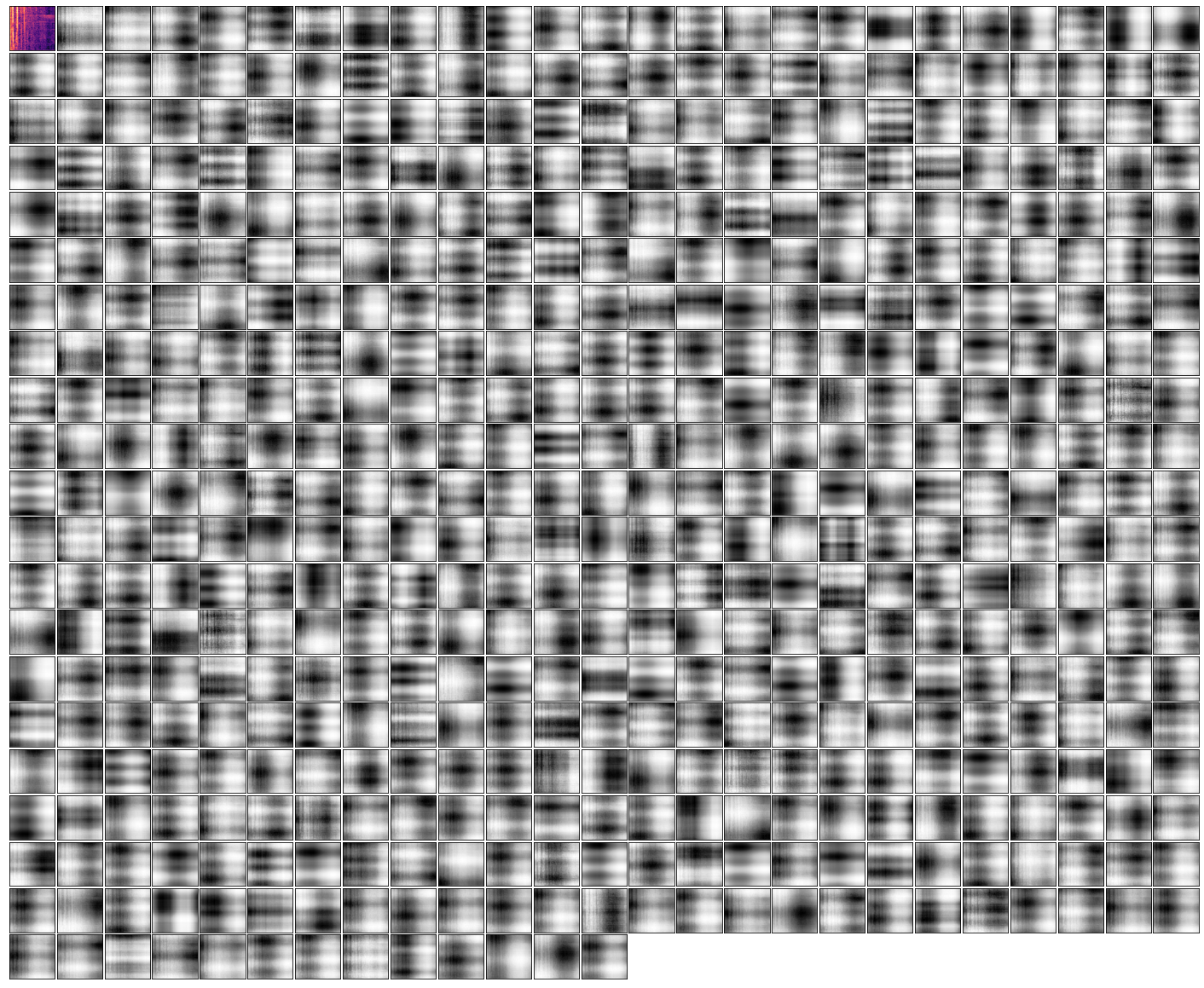}
    \vspace{-12pt}
    \caption{Example attention maps from the \textbf{first cross-attend} over the mel-spectrogram input subset of an AudioSet network trained on \textbf{video and mel-spectrogram} (car).}
\end{figure*}

\begin{figure*}
    \centering
    \includegraphics[keepaspectratio,width=1.0\linewidth]{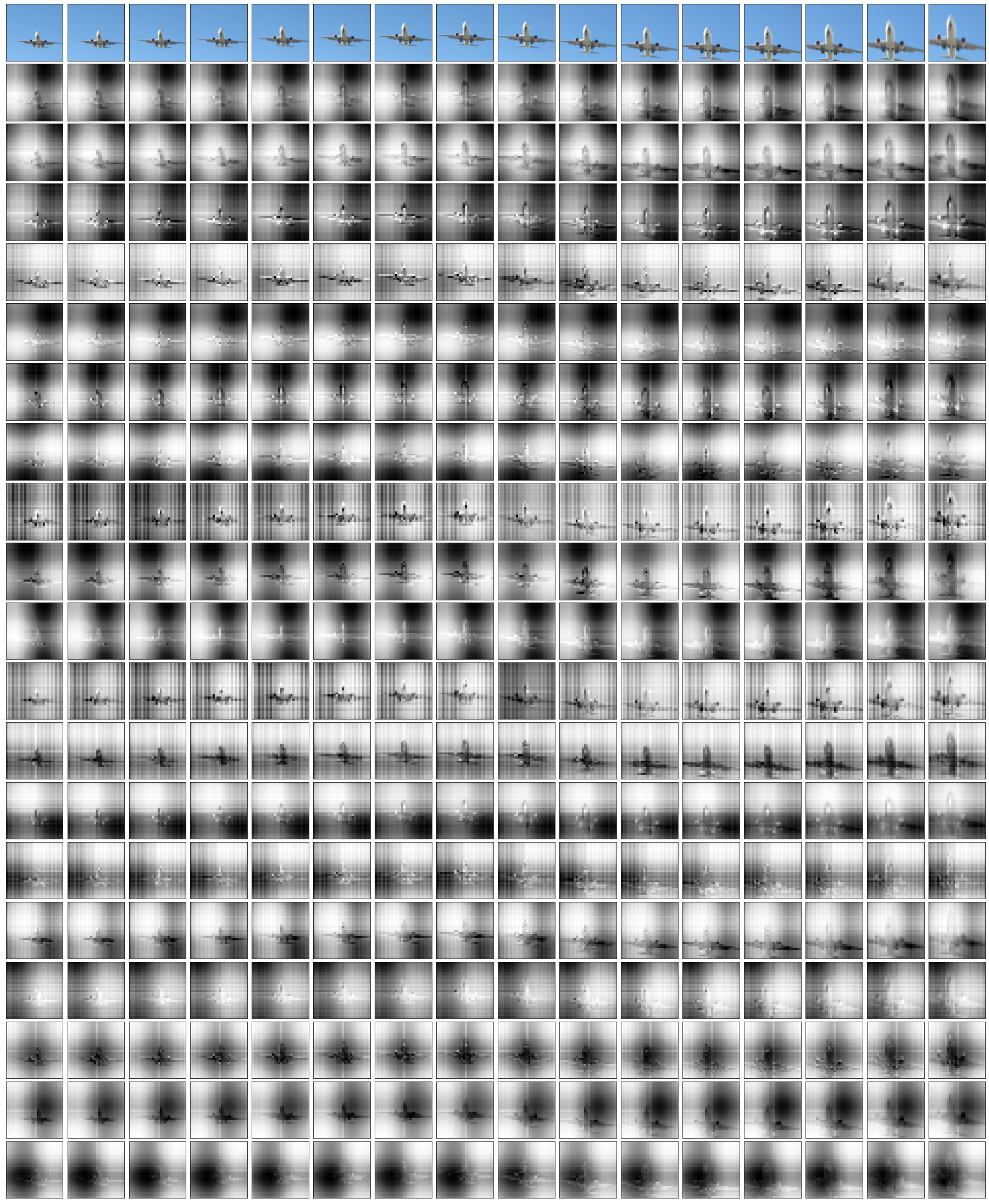}
    \vspace{-12pt}
    \caption{Example attention maps from the \textbf{first cross-attend} over the video input subset of an AudioSet network trained on \textbf{video and mel-spectrogram}.}
\end{figure*}

\begin{figure*}
    \centering
    \includegraphics[keepaspectratio,width=1.0\linewidth]{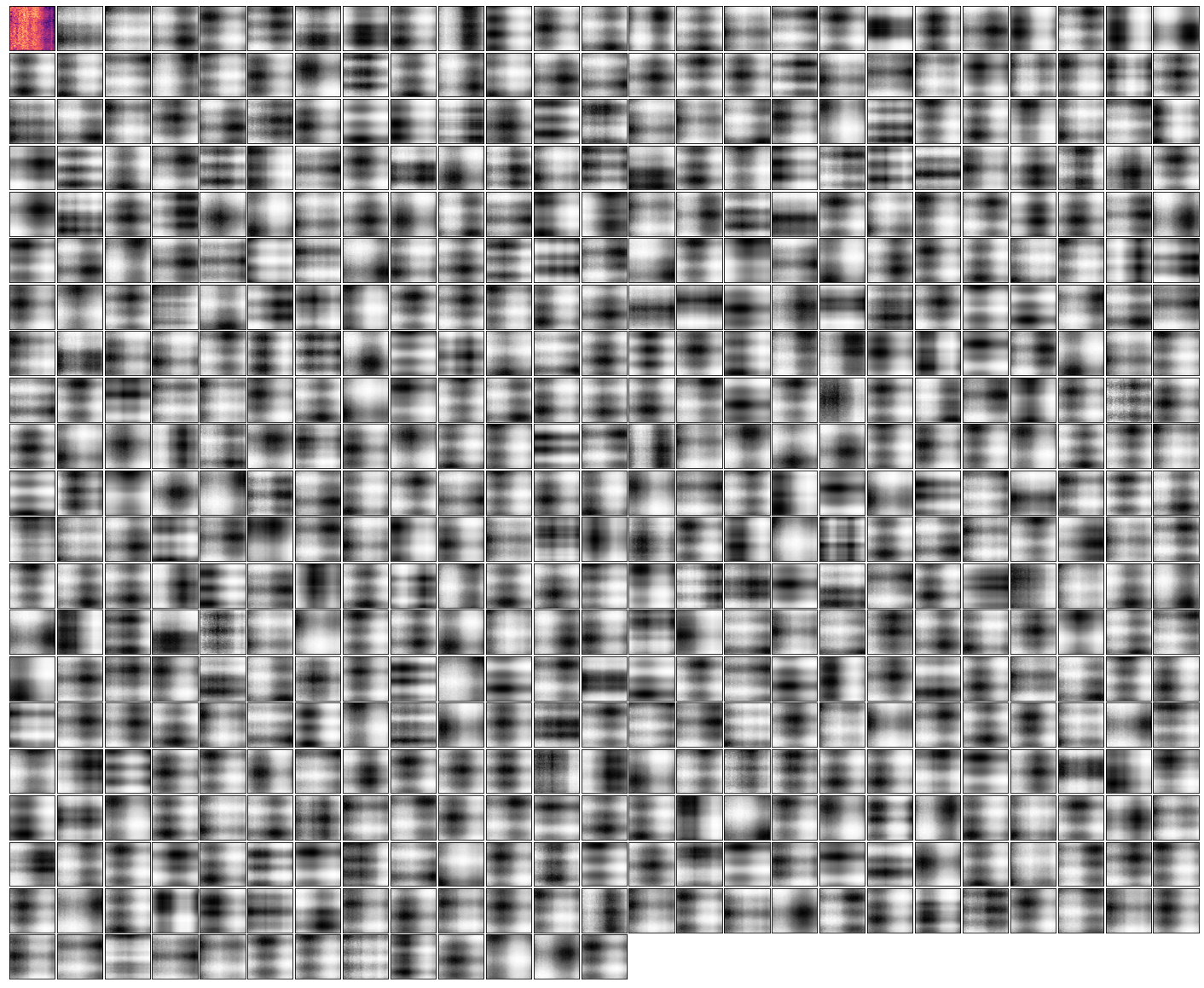}
    \vspace{-12pt}
    \caption{Example attention maps from the \textbf{first cross-attend} over the mel-spectrogram input subset of an AudioSet network trained on \textbf{video and mel-spectrogram} (plane).}
\end{figure*}

\begin{figure*}
    \centering
    \includegraphics[keepaspectratio,width=1.0\linewidth]{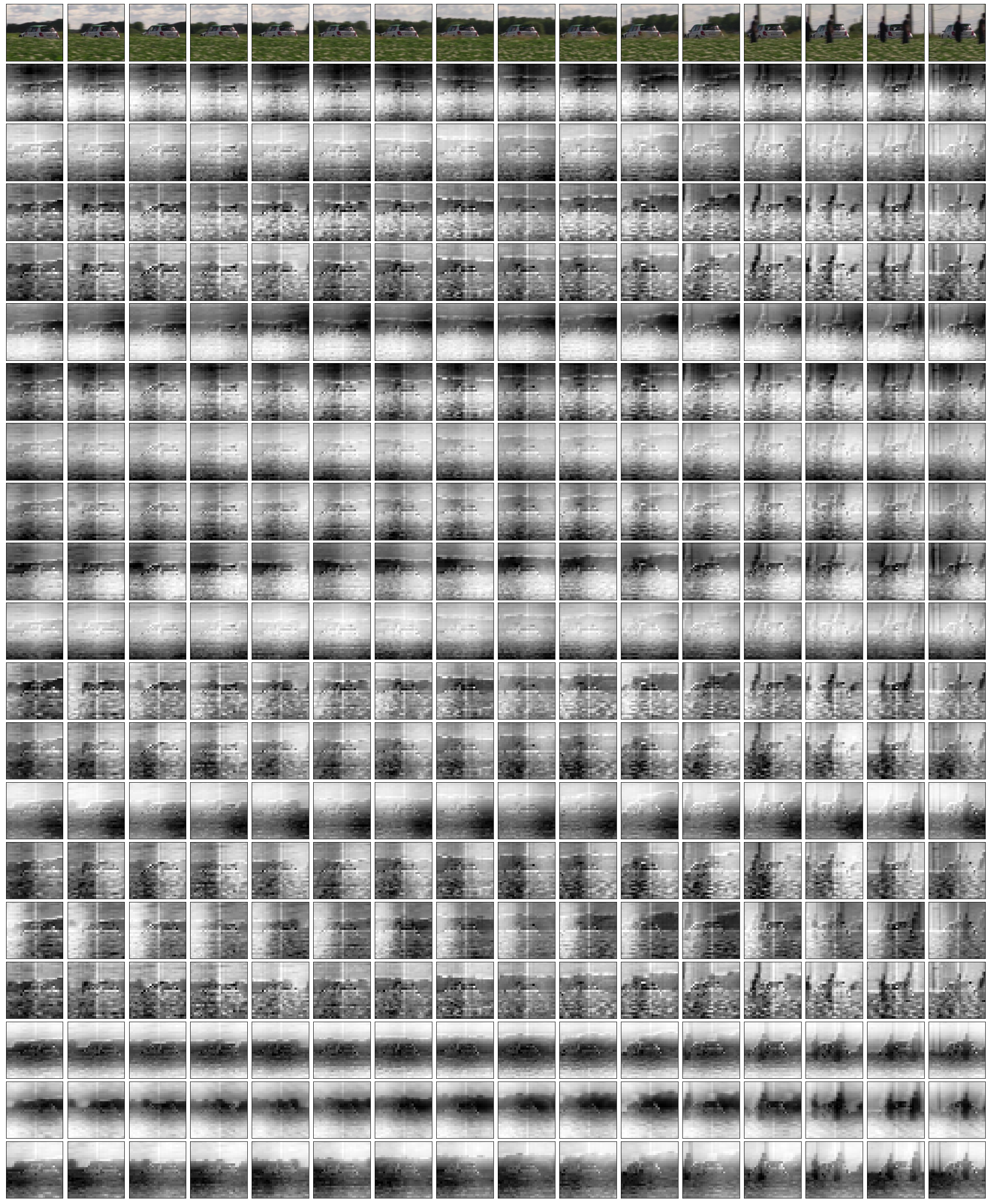}
    \vspace{-12pt}
    \caption{Example attention maps from the \textbf{second (final) cross-attend} over the video input subset of an AudioSet network trained on \textbf{video and mel-spectrogram}.}
\end{figure*}

\begin{figure*}
    \centering
    \includegraphics[keepaspectratio,width=1.0\linewidth]{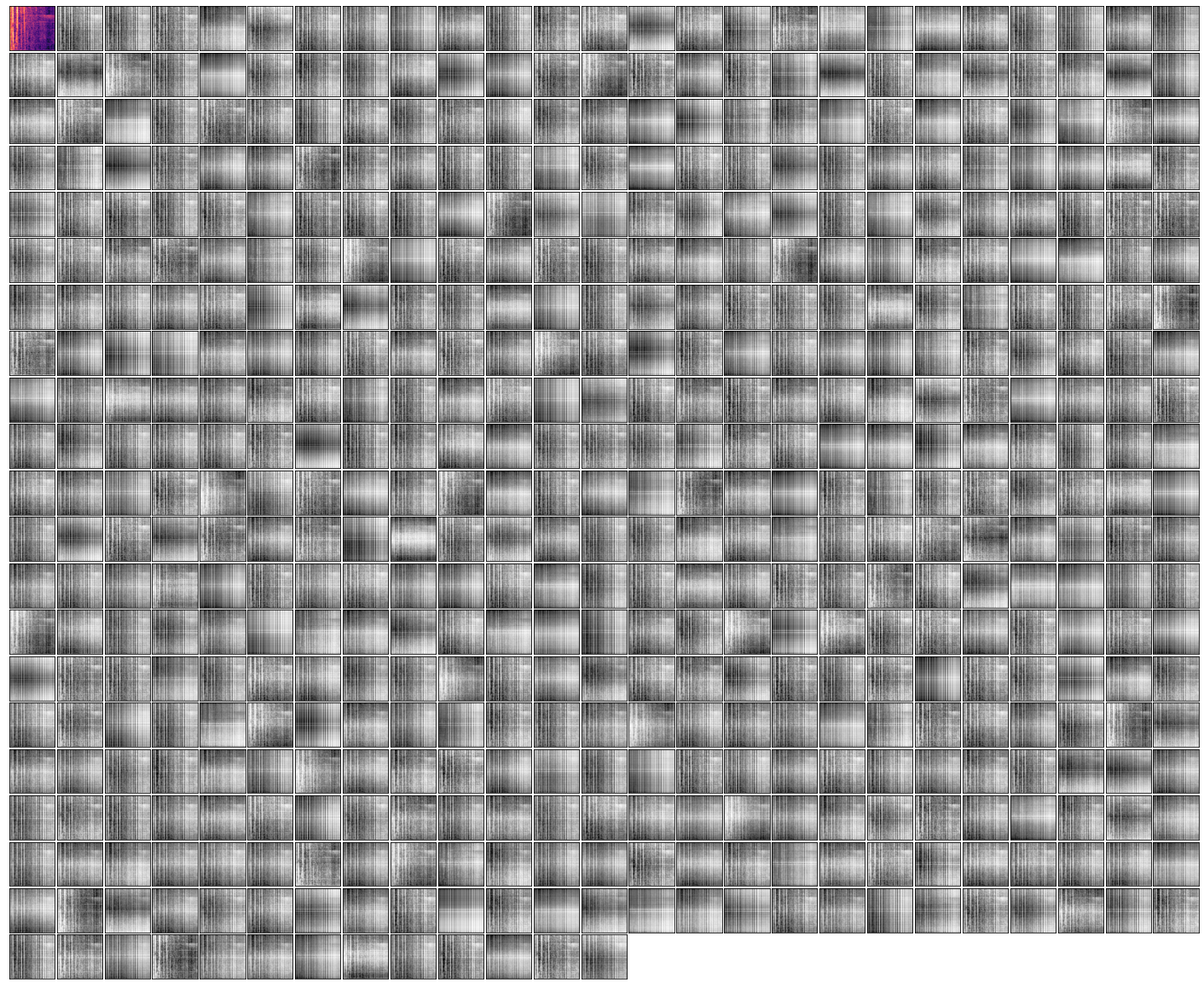}
    \vspace{-12pt}
    \caption{Example attention maps from the \textbf{second (final) cross-attend} over the mel-spectrogram input subset of an AudioSet network trained on \textbf{video and mel-spectrogram} (car).}
\end{figure*}

\begin{figure*}
    \centering
    \includegraphics[keepaspectratio,width=1.0\linewidth]{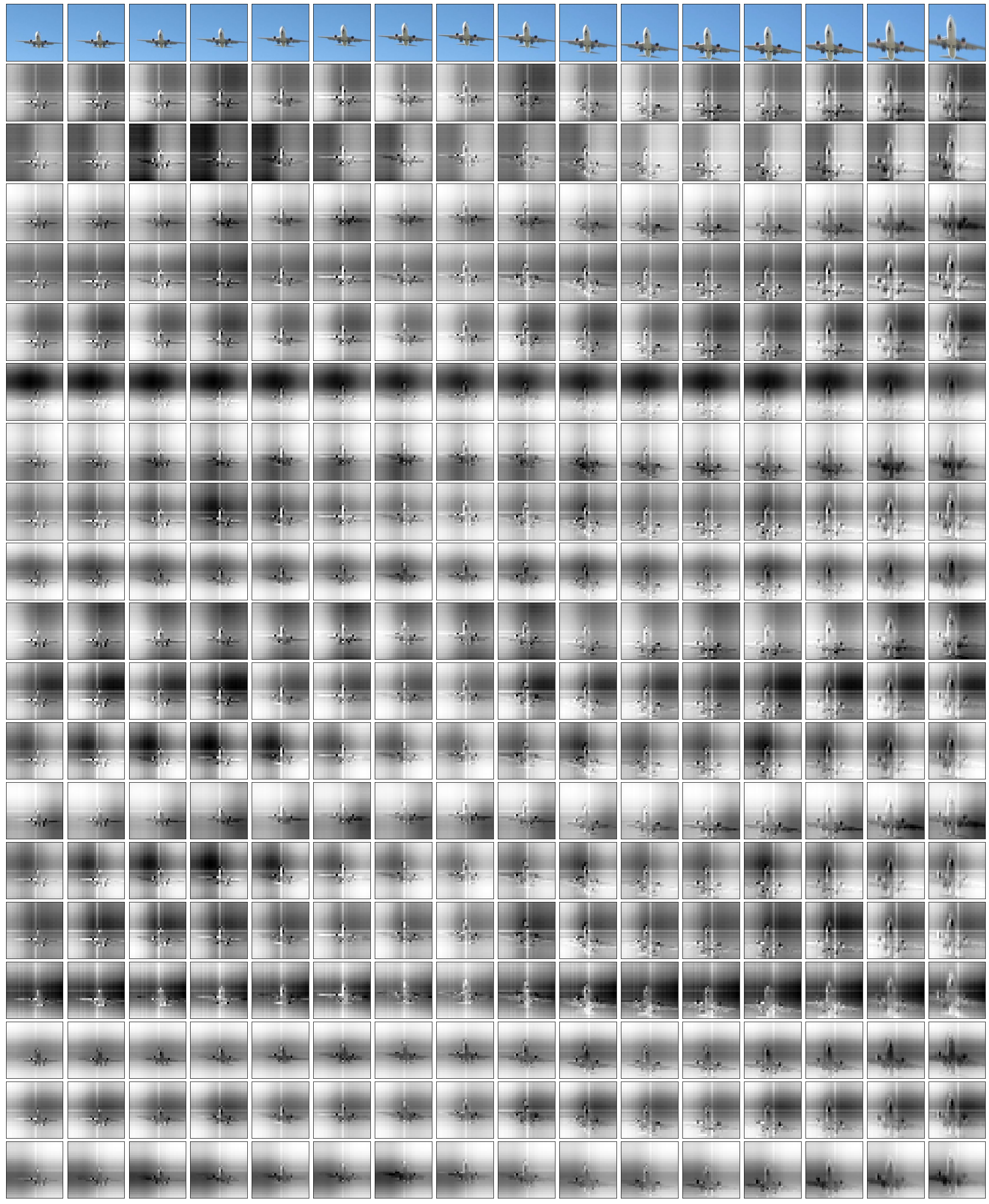}
    \vspace{-12pt}
    \caption{Example attention maps from the \textbf{second (final) cross-attend} over the video input subset of an AudioSet network trained on \textbf{video and mel-spectrogram}.}
\end{figure*}

\begin{figure*}
    \centering
    \includegraphics[keepaspectratio,width=1.0\linewidth]{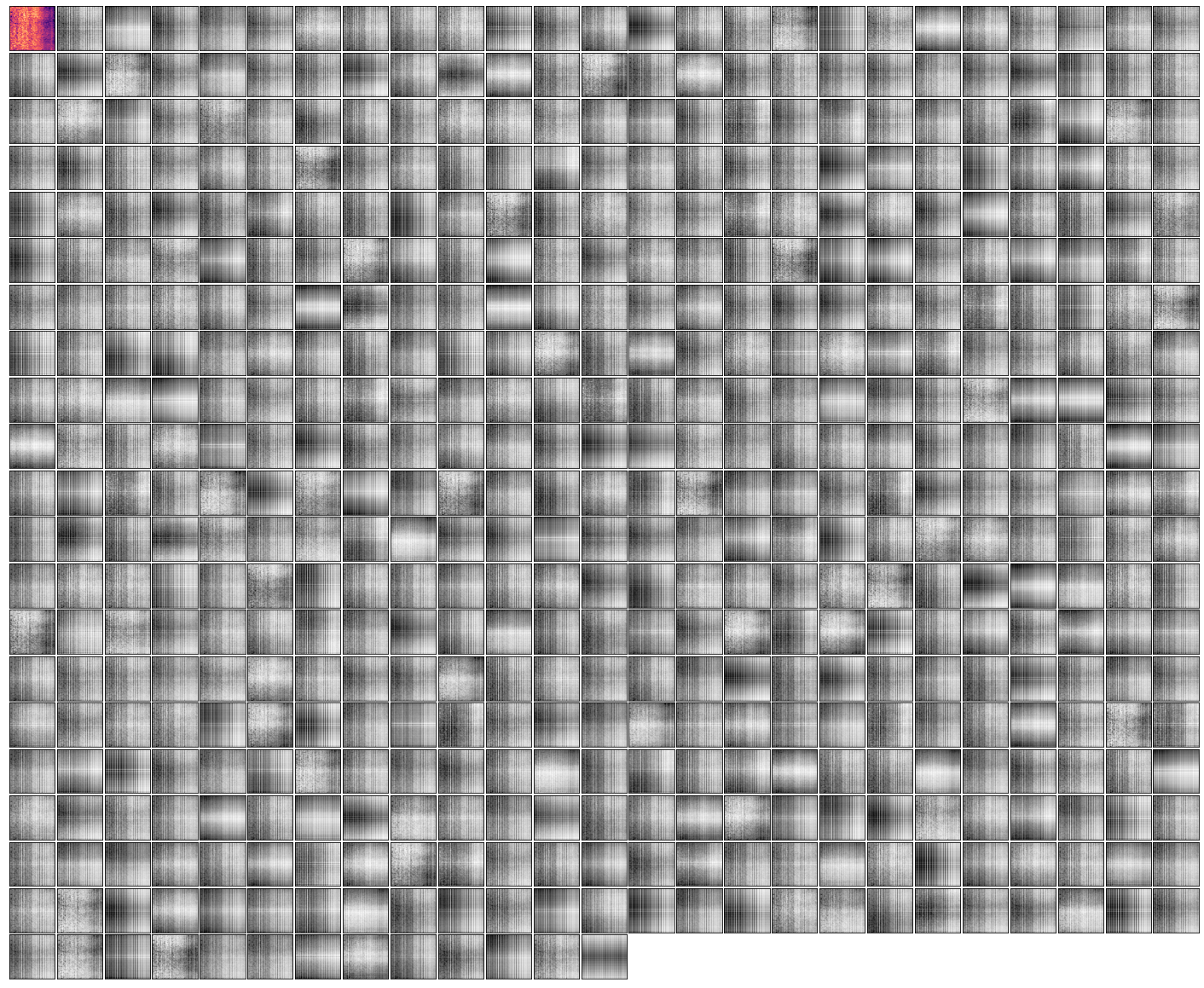}
    \vspace{-12pt}
    \caption{Example attention maps from the \textbf{second (final) cross-attend} over the mel-spectrogram input subset of an AudioSet network trained on \textbf{video and mel-spectrogram} (plane).}
\end{figure*}

\end{document}